\documentclass[letterpaper]{article} % DO NOT CHANGE THIS
\usepackage{aaai24}  % DO NOT CHANGE THIS

\usepackage{times}  % DO NOT CHANGE THIS
\usepackage{helvet}  % DO NOT CHANGE THIS
\usepackage{courier}  % DO NOT CHANGE THIS
\usepackage[hyphens]{url}  % DO NOT CHANGE THIS
\usepackage{graphicx} % DO NOT CHANGE THIS
\usepackage{natbib}  % DO NOT CHANGE THIS AND DO NOT ADD ANY OPTIONS TO IT
\usepackage{caption} % DO NOT CHANGE THIS AND DO NOT ADD ANY OPTIONS TO IT
\usepackage{algorithm}
\usepackage{algorithmic}
\usepackage{booktabs}
\usepackage{tabularray}
\usepackage{multirow}

\usepackage[misc]{ifsym}
\usepackage{twemojis}
\usepackage{url}
\usepackage{amsmath,amssymb}

\usepackage{amsfonts}
\usepackage{helvet}
\usepackage{courier}
\usepackage{url}
\usepackage{float,color}
\usepackage{times}
\usepackage{algorithm}
\usepackage{algorithmic}
\usepackage{float}
\usepackage{pdfpages}
\usepackage{epsfig}
\usepackage{url}
\usepackage{diagbox}
\usepackage{subfigure}
\usepackage{epstopdf}
\usepackage{multicol}
\usepackage{makecell}
\usepackage{longtable}
\usepackage{dsfont,amsfonts,color}
\usepackage{multirow}
\usepackage{booktabs}
\usepackage{xr}
\usepackage{tablefootnote}
\def\0{{\bf 0}}
\def\1{{\bf 1}}

\def\DM{{\mathcal D}}

\def\LM{{\mathcal L}}

\def\MM{{\mathcal M}}

\usepackage{newfloat}
\usepackage{listings}
\DeclareCaptionStyle{ruled}{labelfont=normalfont,labelsep=colon,strut=off} % DO NOT CHANGE THIS
\floatstyle{ruled}
\newfloat{listing}{tb}{lst}{}
\floatname{listing}{Listing}
\title{ConsistentEE: A Consistent and Hardness-Guided Early Exiting Method for Accelerating Language Models Inference}
\author{
    Ziqian Zeng\equalcontrib\textsuperscript{\rm 1}, 
    Yihuai Hong\equalcontrib\textsuperscript{\rm 1}, 
    Hongliang Dai\textsuperscript{\rm 2}, 
    Huiping Zhuang\textsuperscript{\rm 1}, 
    Cen Chen\textsuperscript{\rm 1,3\footnote{Corresponding author.}}
}
\affiliations{
    \textsuperscript{\rm 1}South China University of Technology, China\\
    \textsuperscript{\rm 2}Nanjing University of Aeronautics and Astronautics, China\\
    \textsuperscript{\rm 3}Pazhou Laboratory, China\\
    \{zqzeng, hpzhuang, chencen\}@scut.edu.cn, yihuaihong@gmail.com, hongldai@nuaa.edu.cn
    
}

\usepackage{bibentry}
\begin{document}

\maketitle

\begin{abstract}
Early Exiting is one of the most popular methods to achieve efficient inference. 
Current early exiting methods adopt the (weighted) sum of the cross entropy loss of all internal classifiers as the objective function during training, imposing all these classifiers to predict all instances correctly.
However, during inference, as long as one internal classifier predicts an instance correctly, it can accelerate without losing accuracy. 
Thus, there is a notable gap between training and inference. 
We propose ConsistentEE, an early exiting method that is consistent in training and inference. 
ConsistentEE formulates the early exiting process as a reinforcement learning problem. 
A policy network is added to decide whether an instance should exit or continue. 
The training objective of ConsistentEE only requires each instance to be predicted correctly by one internal classifier. 
Additionally, we introduce the concept \textit{Memorized Layer} to measure the hardness of an instance. 
We incorporate the memorized layer into reward function design, which allows ``easy'' instances to focus more on acceleration while ``hard'' instances to focus more on accuracy. 
Experimental results show that our method outperforms other baselines on various natural language understanding and generation tasks using PLMs and LLMs as backbones respectively. 
\end{abstract}
\section{Introduction}
Recently, pre-trained language models (PLMs) \cite{devlin2019bert, liu2019roberta, yang2019xlnet, brown2020language} and large language models (LLMs) \cite{ouyang2022training} have become fundamental building blocks in the field of natural language processing (NLP).  
As the scales of these models continue to grow, their performance improves but their inference speed slows down. 
This hinders their application in resource-limited scenarios.
To address this problem, many efforts have been made to achieve efficient inference.
There are two lines of work: static and dynamic approaches \cite{zhou2020pabee}. 
Static approaches design lightweight architectures or compress 
the models while dynamic approaches aim to make adaptive inference for each instance. 
Static approaches include weights pruning \cite{michel2019sixteen,voita2019analyzing,fan2020reducing}, quantization \cite{kim2021ibert,yao2022zeroquant,xiao2023smoothquant}, and knowledge distillation \cite{sanh2019distilBERT,sun2019patient,jiao2020tinybert}. 
Dynamic approaches \cite{xu2023survey_dynamic} include token skipping \cite{goyal2020power,kim2021length,ye2021trbert,kim2022learned,guan2022transkimmer,sun2022simple}, and early exiting \cite{zhou2020pabee,xin2021berxit,zhang2022pcee}. 

\begin{figure}[t]
% htbp
\centering

\includegraphics[width=0.474\textwidth]{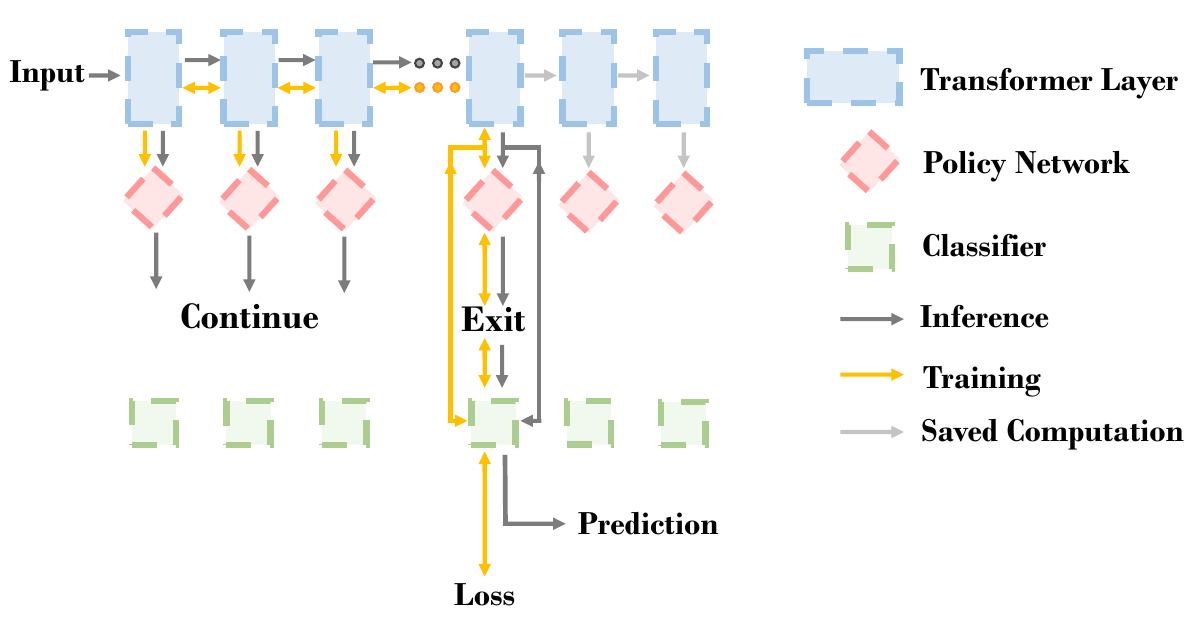}
\caption{The training and inference procedure of ConsistentEE which formulates the early exiting process as a reinforcement learning problem. 
% Besides a internal classifier, a policy network is also added into an intermediate layer. 
A policy network can make two possible actions, i.e., to exit, or to continue. 
If it exits, the corresponding internal classifier is required to predict the instance correctly, otherwise, no loss is incurred by the corresponding internal classifier. 
}
% ConsistentEE model overview. 
\label{fig:main_illustration}

\end{figure}

As one of the most popular methods, early exiting adds an internal classifier to each intermediate layer, allowing instances to stop model inference in an early layer instead of going through the entire model, thus accelerating the inference time.  
Early Exiting can be useful, especially in the era of LLMs. 
% The scale of LLMs will inevitably increase to deal with challenging tasks such as reasoning, planning, etc.
Although the scale of LLMs continues to increase to deal with challenging tasks like reasoning, planning, etc., in the real-world scenario, they may not always encounter such challenging tasks \cite{schuster2022confident}. 
A small sized LLM has been good at handling ``easy'' tasks such as paraphrasing, translation, etc.
When LLMs encounter ``easy'' tasks, early exiting can avoid going through the entire model, thus saving inference time.
% \revisedhl{(reference?) } \revisezq{\cite{schuster2022confident}}

Current early exiting methods typically adopt the (weighted) sum of the cross entropy (CE) loss of all internal classifiers as the training loss, which imposes that all internal classifiers should predict all instances correctly. 
% However, the experimental analysis in Figure \ref{fig:rte_traditional} shows the classification accuracy of each internal classifier is not as satisfactory as expected.
% This observation raises a question: is it necessary for all internal classifiers to predict each instance correctly?
However, our experimental analysis (in Figure \ref{fig:loss_distribution} left) shows that the classification accuracy of each internal classifier is not as satisfactory as expected.
This observation raises a question: is it necessary to require all internal classifiers to predict each instance correctly?
% Actually, in the inference phase, as long as one internal classifier predicts an instance correctly, it can accelerate without sacrificing accuracy.  
Actually, in the inference phase, as long as one internal classifier predicts an instance correctly, the inference can be accelerated without sacrificing accuracy.
Given that there is no such strict requirement during inference, it is reasonable to abandon such a demanding objective during the training phase.

We propose ConsistentEE, an early exiting method that is consistent in training and inference. 
Specifically, ConsistentEE formulates the training process as a reinforcement learning (RL) problem. 
As shown in Figure \ref{fig:main_illustration}, a policy network is introduced at each intermediate layer to determine whether to exit or not. 
If a policy network at an intermediate layer decides to exit, the corresponding internal classifier is expected to predict this instance correctly, otherwise no loss is incurred by this classifier.  
During training, an instance can exit at only one layer. Hence, only one internal classifier is required to predict it correctly. 
As depicted in Figure \ref{fig:loss_distribution} right, unlike existing methods, the ConsistentEE objective enables each layer's classification accuracy to consistently remain at a high level.
% Experimental analysis in Figure \ref{fig:rte_consistent} shows that
% \revisedhl{Our experimental analysis (in Figure \ref{fig:rte_consistent}) shows that unlike the existing method,}
% the classification accuracy of each layer under the ConsistentEE objective is able to maintain a high level. 

% \revisedhl{(Are the accuracies better than the existing method? If not, I don't think it's very necessary to mention this here.)} 
% \revisezq{Yes.}
% \revisezq{Experimental analysis in Figure \ref{fig:rte_consistent} shows that the classification accuracy of each layer under the ConsistentEE objective maintains a high level, significantly higher than that under the traditional objective.}

% The reward function design should consider both accuracy and acceleration. 
% The loss of the internal classifier is consider in our reward function design. 
% Lower losses should obtain higher reward. 
% We involve the depth of the layer at which an instance exits in the reward function. 
% \revisedhl{For the reward function design, we take both accuracy and acceleration into consideration.
% To ensure accuracy, we let the reward function depend on the loss of the internal classifier, and associate lower losses with higher reward values.
% In terms of acceleration, we consider the depth of the layer at which an instance exits.
% }
For the reward function design, we consider both accuracy and acceleration. To ensure accuracy, we incorporate the loss of the internal classifier into the reward function, assigning higher reward values to instances with lower losses. Additionally, to consider acceleration, we involve the layer depth at which an instance exits into the reward function.
Although one can place a trade-off coefficient to balance accuracy and acceleration in the reward function, we argue that instances of different hardness levels should put different weights on accuracy and acceleration.
% ``Easy'' instances generally can be classified correctly at shallow layers. 
We observe that ``easy'' instances generally can be classified correctly at shallow layers.
Those instances should exit as early as possible after they can be classified correctly. 
``Hard'' instances typically can be classified correctly at deep layers.
Those instances should focus on accuracy instead of acceleration at early layers. 
Hence, we also incorporate the hardness level of an instance into the reward function design.
% \revisedhl{(references? if not, mention (here or later) that we have experimental results that can support this claim)} \revisezq{No. Are you suggest to delete the entire passage.}
% \revisedhl{(no, do not delete)}
% Should each instance use the same trade-off coefficient? 
% ``Easy instances'' possibly should focus on acceleration as it is easy to predict them correctly while  ``hard instances'' possibly should pay more attention on accuracy as it is difficult to classify them correctly. 

% How to identify ``easy'' and ``hard'' instances was extensively studied in the literature.  
However, the identification of ``easy'' and ``hard'' instances is itself a difficult problem and is extensively studied in the literature \cite{kumar2010self,arpit2017closer,toneva2018empirical}.
% \revisedhl{(Is it necessary to mention this here?)}.
% \revisedhl{(Deleted.)}
% \cite{kumar2010self,arpit2017closer} use losses at some points during training to measure the hardness of instances \revisedhl{(Is it necessary to mention this here?)}. \cite{toneva2018empirical} came up with the concept ``forgetting events'' to measure the hardness of instances. \cite{toneva2018empirical} consider unforgettable example as ``easy''. Unforgettable examples are defined as examples which are predicted correctly at some point and are persistently correct until the end of training. 
% An example undergo a forgetting event when it was correctly classified at last step but is mis-classified at current step.   
% However, possible values of losses and forgetting events are unbounded. 
% It is not easy to map the loss or forgetting events into a discrete number, i.e., the layer where an instance should exit. 
Inspired by the concept of unforgettable example \cite{toneva2018empirical}, we propose a new concept, \textit{Memorized Layer}, to measure the hardness. 
The memorized layer is the layer where the instance is correctly classified and continuously correctly classified until the final layer. 
Experimental analysis reveals a high and medium correlation between the memorized layer and losses, and forgetting events respectively. 
Our reward function incorporates the ``memorized layer'' to encourage ``easy instances'' to pay more attention to acceleration while ``hard instances'' to focus on accuracy rather than acceleration.

% \cite{kumar2010self} consider an instance with a small loss as ``easy''.
% \cite{arpit2017closer} found that deep networks tend to learn ``easy instances'' first. ``Easy instances'' are defined as instances that correctly classified after one epoch of training. 
% \cite{toneva2018empirical} consider unforgettable examples that are never forgotten once learnt as ``easy instances''. 

To summarize, the contributions of our work are the following:

\begin{itemize}
    \item We propose an early exiting method that can achieve consistency during training and inference by formulating the early exiting problem as a reinforcement learning problem. 
    \item We propose a concept named \textit{Memorized Layer} to measure the hardness of an instance. We incorporate it into the reward function to allow an instance to balance the accuracy and acceleration depending on individual hardness. 
    \item The experimental results show that our method can outperform other baselines on natural language understanding and generation tasks. 
\end{itemize}

Code: \url{https://github.com/ZeroNLP/ConsistentEE}.

\section{Related Work}
\subsection{Early Exiting}
Early exiting methods insert an internal classifier to each intermediate layer, allowing instances to exit at an early classifier rather than at the final classifier. 
According to the exiting criterion, early exiting methods can be categorized into three types: confidence-based or entropy-based, ensemble-based, and learning-based exiting. 
% It achieves acceleration by enabling early termination for easy instances while allowing hard instances to continue processing and undergo further computation. 
% This introduces a trade-off between model accuracy and computational cost. 
% The early exit approach allows for dynamic adjustment of inference acceleration in deep neural networks based on certain features of the input samples, without affecting the model's parameters and scale. 
% According to the training manner of internal classifiers, there are two major types of training objective, i.e., weighted sum of CE losses (SkipBERT, PABEE, Past-Future, PCEE-BERT, LeeBERT), and sum of CE losses (DeeBERT \cite{xin2020deebert}, Right-Tool, Voting, BERxiT, CAT). 
% DeeBERT, FastBERT, RomeBERT, Past-Future use entropy to determine exiting. 
% PABEE, LeeBERT, PCEE-BERT use patient to determine exiting. 
% Right-Tool, SkipBERT, CascadeBERT use calibrated max class probability to determine exit. 

Confidence-based early exiting methods utilize confidence, entropy, or (calibrated) max class probability to exit.  
In DeeBERT \cite{xin2020deebert}, FastBERT \cite{liu2020fastbert}, RomeBERT \cite{geng2021romebert} and Past-Future \cite{liao2021pastfuture}, the instance exits if the entropy is less than a predefined threshold. 
%distilled the final classifier layer to all internal classifiers and used entropy as the exiting criterion. The training objective of FastBERT is the sum of distillation losses. 
% RomeBERT \cite{geng2021romebert} improve FastBERT by using gradient regularization (GR) to facilitate distillation. 
% Past-Future \cite{liao2021pastfuture} incorporates all available past states and future states to make exiting decision. The future states are learned by imitation learners.
%Other early exiting methods such as  also used entropy as the exiting criterion. 
Right-Tool \cite{schwartz2020right}, SkipBERT \cite{wang2022skipbert}, and CascadeBERT \cite{li2021cascadebert} use (calibrated) max class probability as the exiting criterion. 
% In Right-Tool \cite{schwartz2020right}, an instance exits when calibrated maximum class probability is greater than a predefined threshold. 

Ensemble-based early exiting methods utilize predictions from multiple internal classifiers to make better decisions. 
In PABEE \cite{zhou2020pabee}, the instance exits when $k$ consecutive internal classifiers make the same prediction. 
PCEE-BERT \cite{zhang2022pcee} combined both ensemble-based and confidence-based exiting criteria. The instance exits if the confidence scores are greater than a predefined threshold for several consecutive exits.
% LeeBERT \cite{zhu2020leebert} adopted the auto-weighted sum of cross entropy losses and distillation loss as the objective function. 
% Although weights are learnable, it differs from our method because it did not impose that one instance is only predicted at one internal classifier. 
% LeeBERT used the same exiting criterion as PABEE.  

Learning-based methods aim to learn a criterion for early exiting. Our method also falls into this category. 
BERxiT \cite{xin2021berxit} trained a learning-to-exit (LTE) module to predict certainty level, indicating the extent to which the internal classifier can accurately predict the ground truth.
When the output of the LTE is greater than 0.5, an instance can exit. 
CAT \cite{schuster2021cat} introduces a ``meta consistency classifier'' to predict the conformity level, i.e., whether the output of an internal classifier is consistent with the final classifier.
An instance exits when the conformity level is greater than a threshold. 
The policy network in ConsistentEE is different from LTE in BERxit and the meta consistency classifier in CAT in two aspects. 
First, the purposes are different. LTE aims to predict certainty level, and meta consistency classifier aims to predict conformity level, while the policy network determines whether to exit. Second, the ground truth of certainty level and conformity level are available while the ground truth of the layer at which an instance should exit is unknown.

There are two types of training objectives in the above methods, i.e., the weighted sum of cross-entropy losses  and the sum of cross-entropy losses. 
SkipBERT \cite{wang2022skipbert}, PABEE \cite{zhou2020pabee}, Past-Future \cite{liao2021pastfuture}, PCEE-BERT \cite{zhang2022pcee}, and LeeBERT \cite{zhu2020leebert}) used weighted sum of cross entropy losses. 
DeeBERT \cite{xin2020deebert}, Right-Tool \cite{schwartz2020right}, BERxiT \cite{xin2021berxit}, and CAT \cite{schuster2021cat} used sum of cross entropy losses. 
Both objectives require all internal classifiers to predict all instances correctly.

The above methods are layer-wise early exiting methods where the entire input (i.e., all tokens) exits at the same layer. There are some token-wise early exiting methods such as HashEE \cite{sun2022simple} and TR-BERT \cite{ye2021trbert} where different tokens can exit at different layers. 
When a token exits, the attention computation will not be performed on this token at later layers. 
Hence the acceleration comes from the reduced computation cost in the attention module. 
In this sense, token-wise early exiting methods belong to the family of token skipping methods \cite{goyal2020power,kim2021length,kim2022learned,guan2022transkimmer}.

The above methods are early exiting encoding methods to accelerate BERT inference. 
CALM \cite{schuster2022confident} and Free \cite{sangmin2023free} focused on early-exiting for autoregressive models that allocates adaptive computation paths for each token generation. CALM proposed three confidence measures as exiting criteria including Softmax response, hidden-state saturation, and early exit classifier. A calibration procedure is proposed to find a shared exit threshold. 
Free \cite{sangmin2023free} is concurrent with our work.
FREE only allowed two exit points and replaced the state copying mechanism with synchronized parallel decoding to prevent performance degradation.

\section{Methodology}
\subsection{Background and Motivation}
\begin{figure*}[t]
% htbp
\centering
\includegraphics[width=0.8\textwidth]{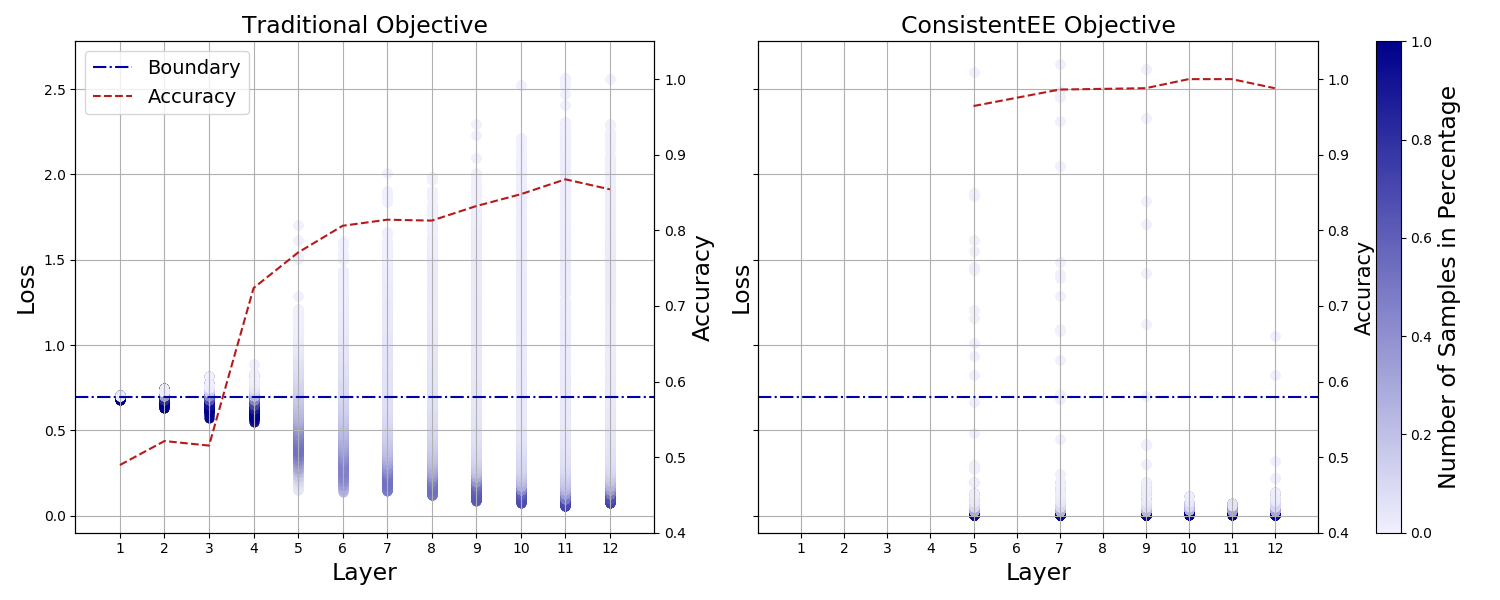}
\vspace{-0.1in}
\caption{Loss values at different layers on the RTE dataset using the weighted sum objective and ConsistentEE objective respectively. The dashed dot line is the classification boundary. A loss above the boundary means misclassification.  
The dashed line is the classification accuracy of each layer. The darker the color, the more samples share the same loss value. }
\label{fig:loss_distribution}
\end{figure*}
% \begin{figure*}[t]
%  \centering
%  \subfigure[Weighted sum objective.]{\label{fig:rte_traditional}
%   \includegraphics[width=0.7\textwidth]{fig/RTE_Traditional.png}
%  }
%  \subfigure[ConsistentEE objective.]{\label{fig:rte_consistent}
%   \includegraphics[width=0.7\textwidth]{fig/RTE_Consistent.png}
%  }
%  \vspace{-0.1in}
%  \caption{Loss values at different layers on the RTE dataset using the weighted sum objective and ConsistentEE objective respectively. The dashed dot line is the classification boundary. A loss above the boundary means misclassification.  
%  The dashed line is the classification accuracy of each layer. The darker the color, the more samples share the same loss value. 
%  } \label{fig:loss_distribution}
% \end{figure*}
The early exiting method adds an internal classifier to each intermediate layer, allowing instances to stop model inference at an early layer without going through the rest of the layers, thus accelerating the inference time. 
% The last exit is the classification layer of the original model.
The traditional training objective function is a weighted sum of the cross-entropy loss of each layer. 
\begin{equation}
\label{eq:sum_of_ce}
    \LM = \sum_{(x,y) \in \DM}\frac{ {\textstyle \sum_{i=1}^{L}}i\cdot l_i }{ {\textstyle \sum_{i=1}^{L}}i} ,
\end{equation}
\begin{equation}
\label{eq:sum_of_ce}
    l_i = H(y,P_i(x)),
\end{equation}
where $(x,y) \in \DM$ is the input-label pair in the dataset, $l_i$ is the cross entropy loss of $i$-th internal classifier, $H(\cdot)$ is the cross entropy function, $y$ is the ground truth distribution represented as a one-hot vector, $P_i(x)$ is the probability distribution produced by the $i$-th internal classifier. 
The earlier internal classifiers have less learning capacity, hence smaller weights are put on 
them \cite{kaya2019shallow,zhou2020pabee}. 
There are also some methods \cite{xin2020deebert,schwartz2020right,xin2021berxit,schuster2021cat} that do not consider varying learning capacities and assign an equal weight to each internal classifier. 
% The weighted sum of cross-entropy losses assigns less weights for shallow classifiers while the sum of cross-entropy loss \cite{xin2020deebert,schwartz2020right,xin2021berxit,schuster2021cat}, assigns an equal weight to each classifier 
% \revisedhl{(This sentence does not connect well with the above part.)}. 

During training, this loss (Eq. \ref{eq:sum_of_ce}) imposes that all internal classifiers should predict all instances correctly. 
% We analyze the loss values of each internal classifiers on the RTE dataset. 
% We also show loss values on other datasets in the Appendix.
% As shown in Figure \ref{fig:rte_traditional}, the performance of the internal classifiers does not behave as expected by traditional training loss. 
% The accuracy of shallow layers is unsatisfactory, especially in the first three layers where it is below 50\%. 
% We guess that the traditional objective is so harsh that the requirement possibly exceeds the limits of capabilities of internal classifiers, and thus harming their performance. 
To investigate whether such an objective is suitable, we analyze the loss values and accuracy of each internal classifier on the RTE dataset. 
%Additional results of some other datasets are included in the Appendix.
As shown in Figure \ref{fig:loss_distribution} (left), with the traditional training loss, the internal classifiers do not perform as expected.
The accuracy of shallow layers is unsatisfactory.
Especially, the first three layers fall below the random guess baseline, i.e., 50\%. 
We think that the traditional objective is too strict that the requirement may have exceeded the limits of the capabilities of internal classifiers, thus harming their performance. 
% which is a binary classification dataset.  
% The training losses are very high under traditional training objective. 

% the burden of each internal classification,
% Many losses have higher value than $-\log 0.5$, which means many instances are predicted incorrectly on MRPC (a binary classification dataset). 

% Experimental analysis shows that it is not possible for all classifiers to correctly classify all samples, which provokes a question whether it is necessary for all internal classifiers to predict each instance correctly?
The experimental analysis provokes a question: whether it is necessary to require all internal classifiers to predict each instance correctly?
Actually, in the inference phase, as long as one internal classifier predicts an instance correctly, it can accelerate without sacrificing accuracy.  
We therefore propose ConsistentEE, an early exiting method that is consistent in training and inference. 
With ConsistentEE, for a training instance, only one internal classifier is required to predict it correctly and this instance will exit at the corresponding layer. 
% As a result, the training process is consistent with the inference process. 
ConsistentEE allows each classifier to focus on correctly classifying partial instances, reducing the burden of each layer. 
As shown in Figure \ref{fig:loss_distribution}, the training losses under ConsistentEE are consistently lower than those under the traditional training objective at every layer. 
Additionally, the accuracy of each layer in the training set remains at a significantly high level. 
The accuracy of layer $1$ to $4$ is not available becauase if the policy network chooses not to exit at these layers, then there is no loss and accuracy in these layers. RTE is a challenging dataset, and no samples exit at these layers.

\subsection{ConsistentEE}
The primary challenge in ConsistentEE is to determine the most appropriate layer for an instance to exit.
As the ground truth exit layer
% of the layer at which instance should exit 
is unavailable, ConsistentEE employs the reinforcement learning (RL) method to automatically learn the optimal layer for an instance to exit during training. 
As shown in Figure \ref{fig:main_illustration}, a policy network is introduced to an intermediate layer besides an internal classifier.
% The input of the policy network and the internal classifier are the same. 
This policy network shares the same input with the internal classifier.
It produces the probability distribution of two actions.
One is exiting at the current layer, and the other one is continuing to the next layer. 
If the policy network takes an action to exit, then the internal classifier at the same layer is required to predict the instance correctly. 
Otherwise, if the policy network takes an action to continue, then there is no loss imposed on the corresponding internal classifier.

The training objective of ConsistentEE involves two parts. 
The first part aims to optimize the policy network so that it can make a good action, i.e., exit or continue. 
The second part aims to optimize the internal classifier so that it can classify the instance correctly. 
Only the internal classifier at the layer at which an instance decides to exit can incur a loss. 
% when the instance can exit at this layer according to the decision of the corresponding policy network. 

Early exiting can be regarded as a sparse reward process because rewards are only received when an instance decides to exit at a particular layer, with no feedback provided for intermediate actions. 
% there is no reward until an instance decides to exit at a particular layer.
% with no feedback provided for intermediate layers. 
% It can be regarded as a sparse reward process because there is no reward until rewards are only received when an instance chooses to exit at a particular layer, with no feedback provided for intermediate layers. 
Considering this characteristic, we use the Policy Gradient technique to automatically discover the optimal exit layer for each instance. We introduce relevant notations in the following.

\textbf{State} $s_t$ is the representation of the input at $t$-th layer. For the classification task with PLMs as backbones, $s_t$ is the representation of \texttt{[CLS]} token at $t$-th layer. 
For the generation task with decoder-only LLMs as backbones, $s_t$ is the hidden state of the last token of the input at $t$-th layer. The hidden state is the output of the last layer (i.e., Add \& Norm layer) in the basic Transformer block.

\textbf{Action} $a_t$ is the action taken at the $t$-th layer. The action space is \{Exit, Continue\}. An instance can exit at the current layer or continue to the next layer.

\textbf{Reward} The reward function should consider accuracy and acceleration. 
For accuracy, we adopt the likelihood of predicting the ground truth as a part of the reward. 
A larger likelihood of predicting the ground truth leads to a larger reward. 
For acceleration, we involve the depth of the layer at which the instance exits.
If an instance exits early (late), then it can gain a larger (smaller) reward. 
The vanilla version of reward $R_{\text{vanilla}}$ is defined as:
\begin{equation}
R_{\text{vanilla}} = \left\{
	\begin{array}{lcl}
	-H(y,P_t(x)) - \alpha \cdot t && {a_t = \text{Exit}} \\
	0 && {a_t= \text{Continue}} 
\end{array} \right.
,
\end{equation}
where $\alpha$ is the trade-off coefficient. 
% where $P_t(x)$ is the probability distribution produced by the internal classifiers at $t$-th layer. 

In the vanilla version of the reward function, each instance has the same trade-off coefficient to balance the accuracy and acceleration.  
% Accuracy and acceleration are two contradictory objectives.  
% Linear combination of them generally cannot reflect the real optimal.
However, ``easy'' instances generally can be correctly classified at shallow layers. 
Such instances should exit as early as possible after they can be correctly classified. 
``Hard'' instances typically can be classified correctly at deep layers.
Such instances should focus on accuracy instead of acceleration at early layers. 
Hence, instances of different hardness levels should put different weights on accuracy and acceleration.
% We introduce a new concept named \textit{Memorized Layer}, denoted as $\MM$, to measure the hardness of an instance. 
% A smaller (larger) value of $\MM$ indicates a easier (harder) instance. 
% To preserve the coherence of our discussion on the reward function, we leave the definition of memorized layer in the next section.
% Then, our reward function that takes instance hardness into consideration is defined as follows,  
Based on this idea, we improve the reward function by taking instance hardness into account:
\begin{equation}
\label{eq:new-reward}
R = \left\{
	\begin{array}{ll}
	-H(y,P_t(x)) - \alpha \cdot (1-\frac{\MM}{L}) \cdot t & {a_t = \text{Exit}} \\
	0 & {a_t= \text{Continue}} 
\end{array} \right.
,
\end{equation}
% where $L$ is the total number of layers. 
% A hard (easy) instance is expected to have a larger (smaller) value of $\MM$, so the trade-off coefficient $\alpha \cdot (1-\frac{\MM}{L})$ is smaller (larger). 
% Hence the reward function put more (less) weight on accuracy compared to the vanilla version. 
where $L$ is the total number of layers; $\MM$ denotes \textit{Memorized Layer}, a new concept we introduced to measure the hardness of an instance. A smaller (larger) value of $\MM$ indicates an easier (harder) instance. More details about this concept will be introduced in the next section. With Eq. \ref{eq:new-reward}, for a hard (easy) instance, the trade-off coefficient $\alpha \cdot (1-\frac{\MM}{L})$ is expected to be smaller (larger), causing it to receive more (less) weight on accuracy compared to the vanilla version.

\textbf{Policy Network} $\pi(a_t|s_t;\theta)$ produces the probability of taking action $a_t$ given the current state $s_t$. 
The Policy network is parameterized by $\theta$. 
% It is a multi-layer perceptron (MLP) in our implementation. 

\textbf{Policy Objective Function}
We optimize the policy network to maximize the expected reward. The policy objective function is defined as:

\begin{equation}
\label{eq:policy}
    J(\theta) = \mathbb{E}_{\tau \sim \pi(a_t|s_t;\theta)} \left[ R(\tau) \cdot  \prod_{t=1}^{T} \pi(a_t|s_t; \theta)  \right],
\end{equation}
where $\tau$ are the trajectories $(s_1,a_1) \cdots (s_T,a_T)$, $T$ is the number of states in the trajectories. Note that $T \leq L$. When $a_T = \text{Exit}$, trajectories terminate at step $T$. There is no further action in the remaining layers. The maximum length of the trajectory is the number of layers. We adopt repeated sampling and $\epsilon$-greedy techniques to allow each instance to have multiple trajectories. 
Eq \ref{eq:policy} is optimized on trajectories generated by all data in the dataset.

\textbf{Task Objective Function}
We optimize the internal classifiers and the backbone to accomplish the task (e.g., classification, generation, etc.) The internal classifiers and the backbone are parameterized by $\omega$. The task objective function is defined as:
\begin{equation}
\label{eq:classifier}
    J(\omega) = \sum_{(x,y) \in \DM}\sum_{t=1}^{T} {H(y,P_t(x)) \cdot \mathds{1}  ( a_t = \text{Exit)}},
\end{equation}
where $\mathds{1}(\cdot)$ is an indicator function that returns $1$ if the condition is satisfied and $0$ otherwise.
% Considering this characteristic, we use Policy Gradient technique, \revisezq{a kind of reinforcement learning method} to allow the model to autonomously discover the optimal exit layer for each sample, and then allocate samples with varying levels of difficulty to the classifiers in each layer based on their respective solving capabilities, aiming to achieve greater accuracy and acceleration as a joint benefit.
 
% Finally, we leverage the mapping capability learned by the Policy network to assign different samples to different layers in order to train the classifiers in a targeted manner. This strategy mitigates the unrealistic expectation of perfect classification for all samples in the original fine-tuning approach, leading to enhanced model performance and effectiveness.

\subsection{Memorized Layer and Hardness of Instance}
Identifying easy and hard instances is the core problem in curriculum learning and has been extensively studied. 
\cite{kumar2010self,arpit2017closer} use losses at some points during training to measure the hardness of instances. \cite{toneva2018empirical} proposed a concept named {unforgettable example} which is predicted correctly at some point and is persistently correct until the end of training. 
{Unforgettable examples} are typically learned in the early stage of training, and may have a low loss during most of the training process. As a result, they may be considered as easy instances according to \cite{kumar2010self,arpit2017closer}.

% However, possible values of losses and forgetting events are unbounded. 
% \revisezq{It is not easy to map the loss or forgetting events into a discrete number, i.e., the layer where an instance should exit. }
% \revisedhl{When involving unbounded variable into the reward function, it is difficult to converge.}

Inspired by the concept of {unforgettable examples} \cite{toneva2018empirical}, we propose a new concept named \textit{Memorized Layer} to measure the hardness of the instance.  
If an instance is correctly classified at a certain layer and remains correctly classified until the final layer, we consider this instance to be successfully memorized at that layer. 
If an instance is memorized at an early (late) layer, we consider it easy (hard).  
We define the layer at which an instance starts to be memorized as {memorized layer}, denoted as $\MM$.
% we consider the instance is successfully memorized at that layer.
% transitions from being classified correctly to incorrectly over the course of learning

\begin{equation}
\label{eq:memorize_layer}
    \MM = k : \forall  i \geq  k, \; \hat{y}_{i} = y,
\end{equation}
where $\hat{y}_i$ is the prediction distribution of $i$-th internal classifier in a one-hot vector form.  In the case that an instance is never predicted correctly, then $\MM = L$, where $L$ is the total number of layers of the backbone. 
Note that $1 \leq \MM \leq L$, which means it is a bounded variable.  
The hardness of the instance can be quantified by the memorized layer. 

% \revisezq{which is a good property because it is more stable in RL training.}

\begin{table}
    \centering
    \begin{tabular}{l|c|c}
        \hline
        Spearman Correlation & Loss & Forgetting Events \\
        \hline
        RTE & 0.76 & 0.59  \\
        % \hline
        MRPC & 0.84 & 0.43  \\
        % \hline
        SST-2 & 0.66 & 0.32  \\
        % \hline
        StackOverflow & 0.74 & 0.56  \\
        \hline
    \end{tabular}
    \caption{Spearman Correlation between memorized layer and loss, and between memorized layer and forgetting events on RTE, MRPC, and SST-2 datasets.}
    \label{tab:correlation}
\end{table}

Table \ref{tab:correlation} shows that Spearman's $\rho$ correlation between {memorized layer} and loss \cite{arpit2017closer}, {memorized layer} and {forgetting events} \cite{toneva2018empirical} on three different datasets. 
An example undergoes a {forgetting event} when it was correctly classified at step $t-1$ but is misclassified at the current step $t$.  
We observe a high correlation between {memorized layer} and loss \cite{arpit2017closer} and a medium correlation between {memorized layer} and {forgetting events}, which indicates that memorized layer generally agrees with other measures of hardness.

\subsection{Model Training and Inference}
During training, we adopt the \textbf{iterative training} technique which iteratively improves the capacity of the policy network and the internal classifiers until convergence is reached. 
% What's more, due to the substantial search space within the reinforcement learning process, we incorporated two supplementary techniques to enhance the effectiveness of exploration by the policy network: \textbf{Repeated Sampling} and \textbf{$\boldsymbol{\epsilon}$-greedy Strategy}. 
% Repeated sampling refers to the practice of obtaining several distinct action sequences for the same sample through the policy network and subsequently calculating their respective rewards. This approach aims to enhance the efficiency of traversing the search space during the exploration process. And the $\boldsymbol{\epsilon}$-greedy strategy states that in the training process of the policy network, there is a probability of $\boldsymbol{\epsilon}$ for randomly deciding whether a sample exits early at a certain layer, while there is a probability of $1-\boldsymbol{\epsilon}$ for the sample's early exit at a certain layer to be determined by the output probability from the policy network. 
% This approach aims to explore a broader range of action sequences. As the number of training steps increases, the value of $\boldsymbol{\epsilon}$ will gradually decrease because of the strengthening capability of the policy network and the growing reliability of its decisions.
The training process is shown as follows:

\begin{enumerate}
    \item Initialization. We adopt the weighted sum of CE losses as the objective function (Eq. \ref{eq:sum_of_ce}) to obtain a good initialization on the internal classifiers and a good estimation on memorized layer. 
    \item Memorized Layer. We calculate memorized layer according to Eq. \ref{eq:memorize_layer}.
   \item Policy Network. We optimize the policy network by maximizing Eq. \ref{eq:policy}. The internal classifiers and the backbone are frozen. 
   \item Internal Classifiers and Backbone. We optimize the internal classifiers and backbone by minimizing Eq. \ref{eq:classifier}. The policy network is frozen. 
   % \item Joint Training. We jointly train the backbone, internal classifiers, and policy network. 
   \item Iterative Training. Repeat steps 2, 3, and 4 convergence is reached.  
\end{enumerate}

During inference, if the probability of taking the action to exit at a particular layer is greater than $0.5$, then the instance will exit at that layer, and we consider the prediction of the internal classifier at that layer as the final prediction.

\section{Experiment}

\begin{table}[!t]
\centering
\begin{tabular}{lrrr}
\toprule
Dataset & Classes & Train/Dev/Test  \\
\midrule
RTE & 2 & 2.5k/0.3k/3.0k \\
MRPC & 2 & 3.7k/0.4k/1.7k \\
SST-2 & 2 & 67k/0.9k/1.8k \\
QNLI & 2 & 105k/5.5k/5.5k \\
QQP & 2 & 364k/40k/391k \\
MNLI & 3 & 393k/9.8k/9.8k \\
\midrule
M-CID	 & 16 & 1.2k/0.2k/ 0.3k \\
StackOverflow & 20 & 18k/1.0k/1.0k \\
\midrule
Alpaca & - &  52k/ \, - \, / \, - \,\\
Dolly & - &  - \, /7.5k/7.5k \\  %  \revisezq{xx} \\
CNN/DM & - & 287k/13.7k/11.5k \\  %  \revisezq{xx} \\
\bottomrule

\end{tabular}
\caption{Statistics of datasets.}
\label{tab: datasets statistics}
\end{table}

\begin{table*}[t]
	\centering
	\scalebox{0.68}{
% 		\begin{tabular}{l|p{0.8cm}<{\centering} p{0.82cm}<{\centering} p{0.8cm}<{\centering} p{0.82cm}<{\centering} p{0.8cm}<{\centering} p{0.8cm}<{\centering} p{0.8cm} }
\begin{tabular}{l|p{0.77cm}<{\centering}p{0.82cm}<{\centering}|*{16}{p{0.82cm}<{\centering}}}

			\toprule
			Method	& \multicolumn{2}{c|}{Averaged} & \multicolumn{2}{c|}{RTE} & \multicolumn{2}{c|}{MRPC} & \multicolumn{2}{c|}{SST-2} & \multicolumn{2}{c|}{QNLI}& \multicolumn{2}{c|}{QQP}& \multicolumn{2}{c|}{MNLI}& \multicolumn{2}{c|}{M-CID} & \multicolumn{2}{c}{Stack.}\\	
   
   & Score & Layer &  Acc & Layer  &  F1 & Layer  &   Acc & Layer &  Acc & Layer & F1 & Layer
            &  Acc & Layer & Acc & Layer & Acc & Layer\\
			\midrule 
			 
            \multicolumn{19}{c}{\textbf{Dev Set}}\\
            \midrule

            BERT-Base 
                & 86.4 & 12 & 66.4 & 12  & 89.6 & 12  & 91.8 & 12 & 90.0 & 12 & 89.6 & 12 & 84.3 & 12 & 89.6 & 12 & 90.2 & 12\\
                \midrule
           % \revisezq{99\% BERT-B}& & \revisezq{65.7}& & \revisezq{88.7}& & \revisezq{90.8}& & \revisezq{89.1} & & \revisezq{88.7} & & \revisezq{83.5}  &  &  \revisezq{88.3} &  & \revisezq{89.3} &   \\ % will delete, just for reference
			% \midrule
                \textbf{Layer-wise} & & & & & & & & & & & &\\
			DeeBERT
			& 85.0 & -29\% &65.7 &-31\%	&88.2 &-35\% &89.6 &-35\% &89.1 &-22\% &87.5 &-45\% &83.3 &-27\% &88.5 &-19\%  &88.2 &-20\%\\	

		    PABEE 
			& 85.1 & -28\% & 65.0 &-18\% & 88.5 & -31\% & 89.8 & -33\% & 89.8 & -18\%  & 88.0 & -35\% & 82.4 & -23\%  &88.6 &-29\%  &88.9 &-35\%\\	
			BERxiT 
			& 85.4 & -35\% & 66.1 & -31\%	&88.4 &-40\% &89.4 &-38\% &90.0 &-23\% & 89.1 & -43\% &83.4 &-35\% &88.6 &-35\%  &88.3 &-33\%\\	
			Right-Tool & 84.8 & -28\%  &65.3 &-17\%	&88.2 &-36\% & 90.1 & -31\%  & 89.4 &-37\%  &  87.9 & -32\%  &  82.9 &-30\% &85.8 &-11\% &88.8 & -30\%\\ 
			PCEE-BERT  & 85.6 & -26\%  &65.5 &-26\%	&88.4 &-17\% &90.5 & -33\% & 89.4 & -38\%  & 88.3 & -33\% &  83.7& -32\% &89.5 &-15\% &89.5&  -15\%
			   \\	
		    \midrule 
			\textbf{Token-wise} & & & & & & & & & & & &\\
			HashEE \twemoji{stopwatch} & 84.3 & -10\% &63.2 &-9\%  &85.6 &-10\%  &90.3 & -4\% & 88.4 & -10\% & 85.5 & -6\%  & 82.6 & -19\%   &89.3 & -7\% & 89.3 & -15\% \\
   
			TR-BERT \twemoji{stopwatch} & 85.1 & -10\% &65.5 &-6\%  &87.7 & -5\%   &90.6 & -6\%  & 88.6 & -11\% & 86.3 & -9\%  & 83.1 & -15\%    &89.4 & -4\%   & 89.5 & -20\%    \\
            \midrule
            \textbf{ConsistentEE} \twemoji{stopwatch} &\textbf{85.8} &\textbf{-35\%} & 65.7 & -39\%  &88.6 & -32\%   &90.7 & -36\% &89.3  & -34\%  & 89.2 & -43\%  &83.6  & -30\%   &89.5 & -26\% &  89.8  &  -37\%  \\ 
			\textbf{ConsistentEE}
			&\textbf{85.8} &\textbf{-51\%} &\textbf{65.7} &\textbf{-55}\%  &\textbf{88.6} &\textbf{-53\%}  &\textbf{90.7} &\textbf{-58\%} &\textbf{89.3}&\textbf{-57\%} &\textbf{89.2} & \textbf{-55\%} &\textbf{83.6} &\textbf{-39\%}   &\textbf{89.5}  &\textbf{-41\%}  &\textbf{89.8}  &\textbf{-48\% }\\

            % &\textbf{66.8} &\textbf{-28\%}
			\textbf{ConsistentEE} $\star$
			& 86.9 & -34\% &\textbf{66.4} &\textbf{-44}\%  &89.9 & -30\% & 92.0 & -35\% & \textbf{90.4} & \textbf{-42\%} & 90.2 & -43\%  & 84.6 &-27\%  & 90.9  & -11\% & \textbf{90.7}  &\textbf{-42\%} \\
                \midrule
                \multicolumn{19}{c}{\textbf{Test Set}}\\
                \midrule
                BERT-Base & 87.0 & 12 & 69.1& 12 & 88.9 & 12 & 93.1 & 12 & 89.8 & 12 & 89.2 & 12 & 83.2 & 12 & 91.7 & 12 & 91.0 & 12
                \\
                PABEE & 85.7 & -19\% &  67.6 & -23\% &88.0 &-17\% & 90.5  & -24\% &  89.1  & -18\% & 88.4 & -22\% & 81.4 & -12\% & 90.1 & -20\% & 90.4 & -13\% \\
                BERxiT & 85.8 &-20\% &  67.9 & -21\% &87.8 &-14\%  & 90.3  & -27\% & 89.4  & -20\% & 88.7 & -16\% & 82.4  & -14\% & 89.9 &  -19\% &90.0 & -29\% \\
                DeeBERT & 85.2 &-18\% &  67.2 & -22\% &87.3 &-11\%  & 89.9  & -26\% & 88.6  & -20\% & 88.2 & -13\% & 82.0  & -9\% & 89.3 &  -17\% &89.1 & -27\%
                \\
                \textbf{ConsistentEE}
			& \textbf{87.1} & \textbf{-41\%} & \textbf{69.0} & \textbf{-46\%} & \textbf{89.0} & \textbf{-37\%} & \textbf{92.9} & \textbf{-46\%} & \textbf{89.9} & \textbf{-42\%} & \textbf{89.0} & \textbf{-45\%} & \textbf{83.4} & \textbf{-31\%} & \textbf{91.9} & \textbf{-45\%} & \textbf{91.4} & \textbf{-36\%}  \\
                
			\bottomrule
    	\end{tabular}}
 	\caption{Comparison among ConsistentEE and baselines on eight datasets. The evaluation metrics for model quality are accuracy or F1 scores. The evaluation metrics for model acceleration are saved layers/runtime (w.r.t BERT-Base). Methods using saved runtimes as metrics are marked with \twemoji{stopwatch}. Methods with no accuracy loss are marked with $\star$. Results which exhibit better accuracy and saved layers than other methods are highlighted in \textbf{bold}.
    }
    % Averaged scores and saved layers/runtime on eight datasets are reported. 
    \label{tab:main_result}
\end{table*} 

\begin{figure*}[!t]
% htbp
\centering

\includegraphics[width=0.9\textwidth]{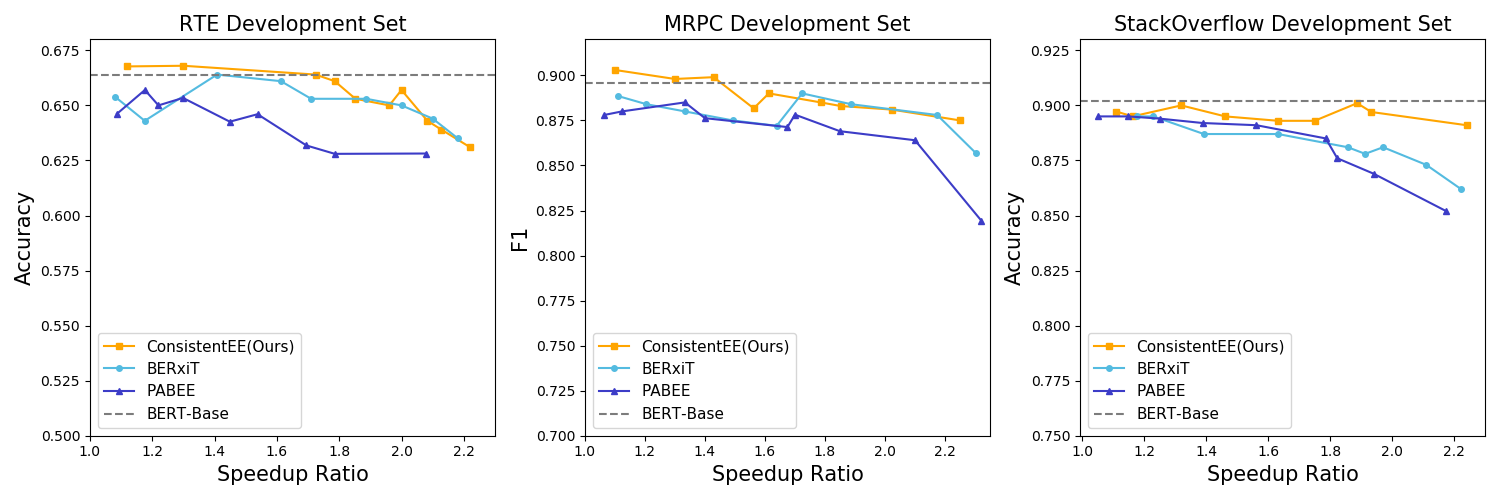}
\caption{Accuracy-Speed curves on the BERT-Base model. The evaluation metric for speedup is saved layers.}
% \caption{Model quality and efficiency trade-offs on the BERT-Base model. The evaluation metrics for model quality and efficiency are accuracy and saved layers respectively.}

\label{fig: speed ratio on 3 models}

\end{figure*}
\subsection{Experimental Settings}
To evaluate acceleration capacities on the classification task with PLMs as backbones, we conduct experiments on six classification datasets of the GLUE benchmark \cite{wang2018glue} and two multi-classes classification datasets including M-CID \cite{arora2020cross} and StackOverflow \cite{xu2015stackoverflow}. 
M-CID and StackOverflow are intent classification datasets which are collected from Covid-19 services and online questions. 
Since most GLUE datasets are binary classification datasets, it is possible that instances may still be classified correctly despite experiencing premature or hasty exits. Hence we evaluate early exiting methods on some multi-classes classification datasets. 
Statistics of these datasets are listed in Table \ref{tab: datasets statistics}. 
We compared our method with various baselines including DeeBERT\cite{xin2020deebert}, PABEE \cite{zhou2020pabee}, BERxiT \cite{xin2021berxit}, Right-Tool \cite{schwartz2020right}, PCEE-BERT \cite{zhang2022pcee}, HashEE \cite{sun2022simple}, and TR-BERT \cite{ye2021trbert}.
For layer-wise exiting methods, we use saved layers to evaluate model acceleration. 
According to \cite{xin2020deebert}, saved layers are linearly proportional to actual runtime. 
While for token-wise exiting methods, we use saved runtime. 
% For baselines, we run the codes provided by their original papers and report results. 
Most results are reported on the development set, since the large number of evaluations are restricted by the GLUE evaluation server. 
% Most results are reported on the development set.
Only BERT-Base \cite{devlin2019bert}, three competitive baselines, and ConsistentEE are evaluated on the test set.

To evaluate acceleration capacities on the generation task with LLMs as backbones, we perform the supervised fine-tuning step using Alpaca \cite{taori2023alpaca} as training data and test on the Dolly dataset \cite{conover2023DollyV2}. We also evaluate our method on text summarization task on CNN/DM dataset \cite{ramesh2016cnn}. The backbone LLMs are LLaMA-7b and LLaMA-13b \cite{hugo2023llama}. Due to our limited computational resources, we train the LLM backbone with LoRA \cite{edwardlora2021} in the initialization step. Subsequently, we freeze the LLM backbone and only train the internal classifiers and policy networks. To save pre-processing time, we adopt the vanilla reward function rather than the hardness-guided reward function because the time complexity of computing the memorized layer is linear to the number of tokens in generated responses.
We set the beam to 1, top-sampling to 0.75, top-k selection to 40 and temperature to 1.
The baseline is CALM \cite{schuster2022confident}. We use Rouge-L \cite{lin-2004-rouge} and BERT-F \cite{zhang2019bertscore} scores to measure the model quality and use save layers to measure the model acceleration. When we compute Rouge-L and BERT-F, we use the response without acceleration as the reference on Alpaca/Dolly dataset and use the gold summary as the reference on CNN/DM.
Following \cite{elbayad2020depth,schuster2022confident}, we use a state copying strategy to handle the computation of hidden states.
Specifically, the computation of the input hidden state of token $t$ at layer $i$ relies on the output hidden states of the previous layer $i-1$ for all preceding tokens $\{1, \cdots, t-1\}$. In cases where some preceding tokens exit before the $i-1$ layer, we copy the output hidden state at the layer at which the token exits, and use it as the output hidden state in $i-1$ layer.

We use a two-layer Multi-Layer Perceptron (MLP) to implement the policy network. 

\subsection{Main Results on Classification}

We evaluate the inference capabilities of ConsistentEE and baselines on the eight datasets.
The result of comparisons are shown in Table \ref{tab:main_result}. 
On the development set, with no accuracy loss, the averaged saved layers of ConsistentEE is 34\%, which shows that our method can achieve 1.54x acceleration without sacrificing accuracy.  
If a little loss (1\% w.r.t the performance of BERT-Base) of accuracy is tolerated, the averaged saved layers of ConsistentEE is 51\%, which outperforms the best baseline BERxiT \cite{xin2021berxit} by 16\%. 
ConsistentEE surpasses token-wise baselines by 25\% in terms of runtime performance. 
On the test set, ConsistentEE also outperforms BERxiT. 
% In addition, the saved layers of ConsistentEE are consistently larger than the best baseline on eight datasets.
% However, the best baselines on eight datasets are varied. 
% BERxiT \cite{xin2021berxit}
% is the best baseline on 5 datasets. PCEE-BERT \cite{zhang2022pcee}, PABEE \cite{zhou2020pabee}, and RightTool \cite{schwartz2020right} are the best baselines on QNLI, M-CID, and StackOverflow respectively. 

Figure \ref{fig: speed ratio on 3 models} shows the accuracy-speed curves of ConsistentEE, PABEE, and BERxiT on three datasets. 
Notably, ConsistentEE mostly has higher accuracy than PABEE and BERxiT under different speed-up ratio, showing its superiority over the baselines.
Furthermore, we have observed that when the speedup ratio is not overly demanding, the accuracy of ConsistentEE surpasses that of BERT-base. This finding suggests that ConsistentEE is a cautious method that prioritizes accuracy over acceleration. It ensures high accuracy while also offering the potential for speed improvements.

\subsection{Ablation Study}
\begin{figure}[!t]
% htbp
\centering

\includegraphics[width=0.40\textwidth]{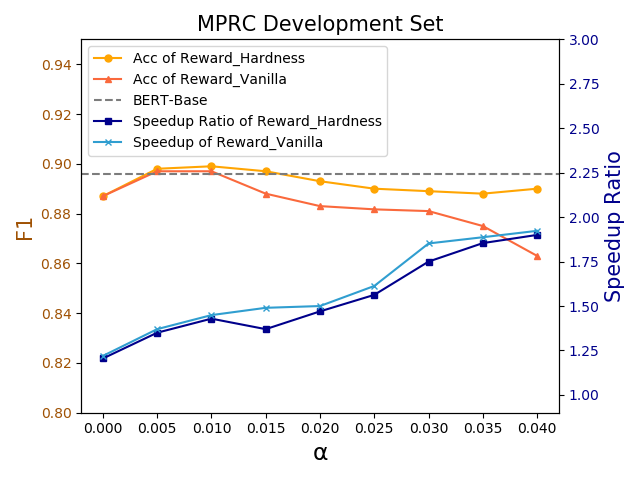}
\caption{Accuracies and speedup ratios of different reward functions under varied $\alpha$.}

\label{fig:MRPC_ablation}

\end{figure}

As depicted in Figure \ref{fig:MRPC_ablation}, the performance of the vanilla reward and the hardness-guided reward are comparable when the speedup ratio is small. However, as the speedup ratio increases, the performance of the vanilla reward deteriorates significantly, while the hardness-guided reward maintains a satisfactory level of accuracy. The possible reason is that the vanilla reward treats each instance equally, causing hard instances struggle to maintain accuracy as they prioritize acceleration as easy instances do. 
In contrast, the hardness-guided reward encourages instances to assign different weights to accuracy and acceleration based on their individual hardness. 
Simple samples can be correctly classified effortlessly and achieve as much acceleration as possible. On the other hand, hard instances place a greater emphasis on accuracy while paying less attention to acceleration, which is more likely to achieve a satisfactory level of accuracy. This differentiated treatment allows the hardness-guided reward to effectively balance the trade-off between accuracy and acceleration across various instances.

\subsection{Hyperparameter Analysis}

\begin{figure}[!t]
% htbp
\centering

\includegraphics[width=0.40\textwidth]{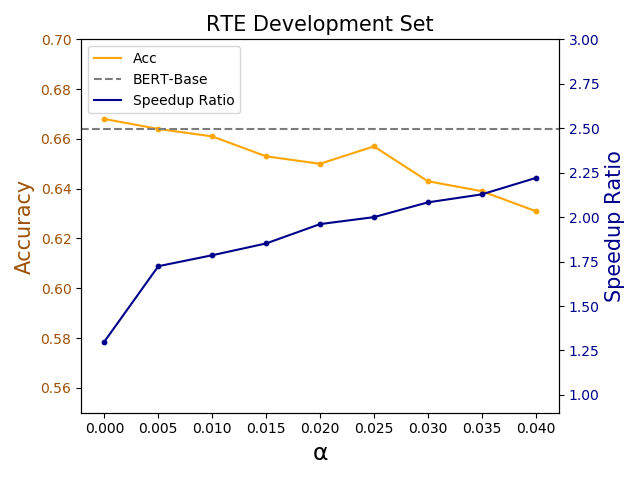}
\caption{Accuracies and speedup ratios of ConsistentEE under varied $\alpha$.}

\label{fig:hyperparameter}

\end{figure}

The search range of $\alpha$ is $\{0.0, 0.005, 0.010, \cdots, 0.04\}$. 
The small value of $\alpha$ is chosen because the CE loss is approximately two orders of magnitude smaller than $t$ (the depth of the exit layer). As shown in Figure \ref{fig:hyperparameter}, a larger value of $\alpha$ leads to a higher speedup ratio and a larger loss in accuracy. 
The accuracy exhibits a variation of only 4\% across the entire search range, indicating that ConsistentEE is relatively insensitive to $\alpha$.

\subsection{Main Results on Generation}
\begin{figure}[t]
% htbp

\centering

\includegraphics[width=0.478\textwidth]{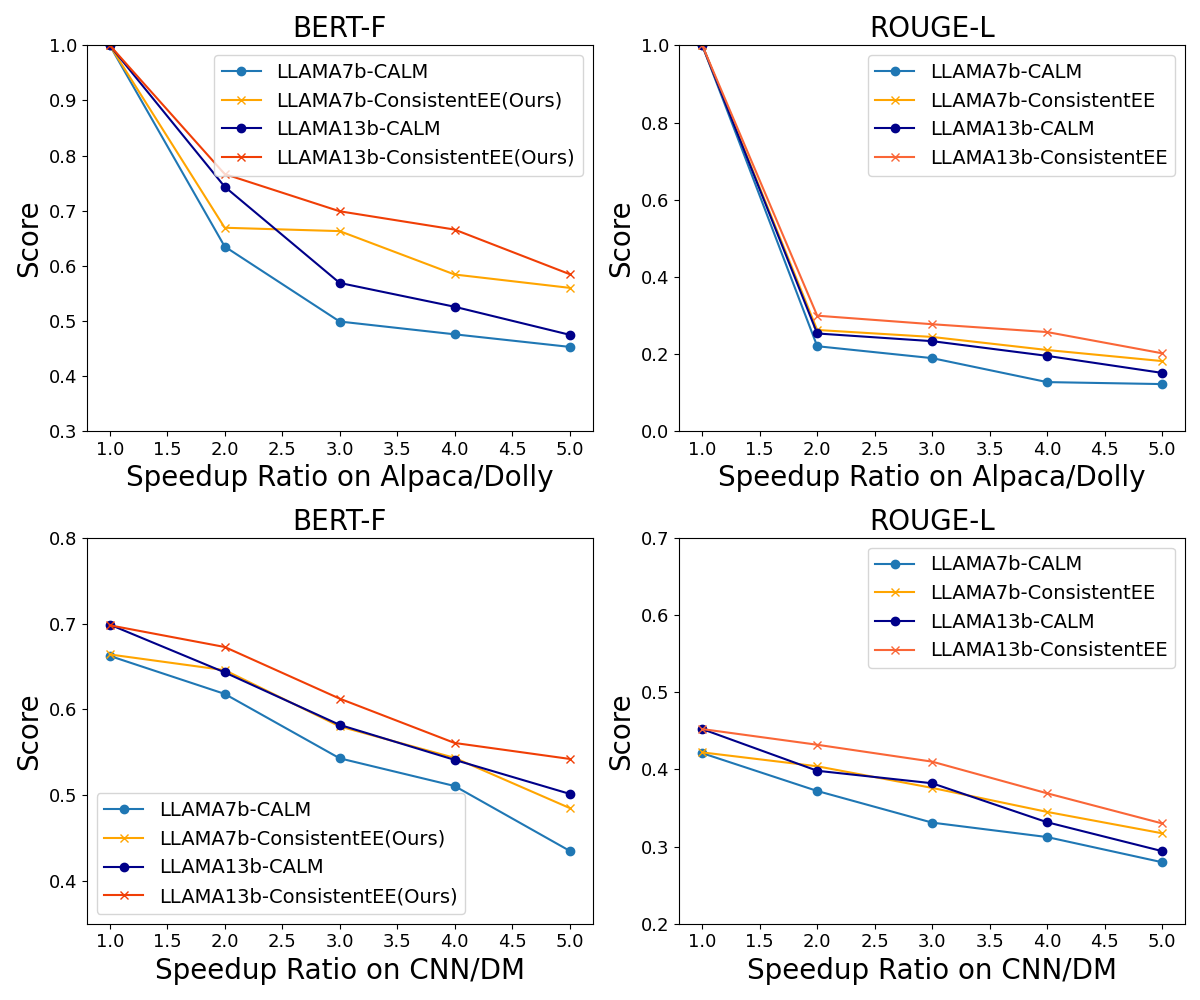}
\caption{BERT-F and Rouge-L and speedup ratios of CALM and ConsistentEE.}

\label{fig:speed up on LLM}

\end{figure}

As shown in Figure \ref{fig:speed up on LLM}, on the Alpaca/Dolly dataset, ConsistentEE and CALM demonstrate similar performance when the speedup ratio is below 2x. However, as the speedup ratio increases, ConsistentEE outperforms CALM. 
On the CNN/DM dataset, ConsistentEE outperforms CALM consistently under different speedup ratios.
Examples of generation from ConsistentEE under different speedup ratios result are shown in Table \ref{tab:case_study} in Appendix. 
The responses generated under 2.0x appear reasonable and closely resemble the original response. However, the responses generated under 5.0x vary from the original response but still make sense. 

\section{Conclusion}
We propose a reinforcement learning based approach to early exiting, so that at the training phase, only one internal classifier is required to predict the instance correctly. This makes the training phase consistent with the inference phase and can allow each layer to obtain better classification accuracy. 
For the reward function of the reinforcement learning framework, we propose the concept \textit{memorized layer} to measure the hardness of each instance, and use it to dynamically balance accuracy and acceleration instead of using a fixed coefficient.
Experimental results conducted on various datasets show that our approach is able to outperform the competitive baselines, demonstrating its effectiveness and efficiency.

\section*{Acknowledgement}
This research was supported by the Guangzhou Basic and Applied Basic Research Foundation (Grant No. 2023A04J1687), National Natural Science Foundation of China (Grant No. 6230070401), South China University of Technology-TCL Technology Innovation Fund.

\bibliography{aaai24}

\begin{thebibliography}{50}
\providecommand{\natexlab}[1]{#1}

\bibitem[{Arora et~al.(2020)Arora, Shrivastava, Mohit, Lecanda, and Aly}]{arora2020cross}
Arora, A.; Shrivastava, A.; Mohit, M.; Lecanda, L. S.-M.; and Aly, A. 2020.
\newblock Cross-lingual transfer learning for intent detection of covid-19 utterances.

\bibitem[{Arpit et~al.(2017)Arpit, Jastrzebski, Ballas, Krueger, Bengio, Kanwal, Maharaj, Fischer, Courville, Bengio et~al.}]{arpit2017closer}
Arpit, D.; Jastrzebski, S.; Ballas, N.; Krueger, D.; Bengio, E.; Kanwal, M.~S.; Maharaj, T.; Fischer, A.; Courville, A.; Bengio, Y.; et~al. 2017.
\newblock A Closer Look at Memorization in Deep Networks.
\newblock \emph{stat}, 1050: 1.

\bibitem[{Bae et~al.(2023)Bae, Ko, Song, and Yun}]{sangmin2023free}
Bae, S.; Ko, J.; Song, H.; and Yun, S. 2023.
\newblock Fast and Robust Early-Exiting Framework for Autoregressive Language Models with Synchronized Parallel Decoding.
\newblock In \emph{EMNLP}, 5910--5924.

\bibitem[{Brown et~al.(2020)Brown, Mann, Ryder, Subbiah, Kaplan, Dhariwal, Neelakantan, Shyam, Sastry, Askell, Agarwal, Herbert{-}Voss, Krueger, Henighan, Child, Ramesh, Ziegler, Wu, Winter, Hesse, Chen, Sigler, Litwin, Gray, Chess, Clark, Berner, McCandlish, Radford, Sutskever, and Amodei}]{brown2020language}
Brown, T.~B.; Mann, B.; Ryder, N.; Subbiah, M.; Kaplan, J.; Dhariwal, P.; Neelakantan, A.; Shyam, P.; Sastry, G.; Askell, A.; Agarwal, S.; Herbert{-}Voss, A.; Krueger, G.; Henighan, T.; Child, R.; Ramesh, A.; Ziegler, D.~M.; Wu, J.; Winter, C.; Hesse, C.; Chen, M.; Sigler, E.; Litwin, M.; Gray, S.; Chess, B.; Clark, J.; Berner, C.; McCandlish, S.; Radford, A.; Sutskever, I.; and Amodei, D. 2020.
\newblock Language Models are Few-Shot Learners.
\newblock In \emph{NeurIPS}.

\bibitem[{Conover et~al.(2023)Conover, Hayes, Mathur, Xie, Wan, Shah, Ghodsi, Wendell, Zaharia, and Xin}]{conover2023DollyV2}
Conover, M.; Hayes, M.; Mathur, A.; Xie, J.; Wan, J.; Shah, S.; Ghodsi, A.; Wendell, P.; Zaharia, M.; and Xin, R. 2023.
\newblock Free Dolly: Introducing the World's First Truly Open Instruction-Tuned LLM.

\bibitem[{Devlin et~al.(2019)Devlin, Chang, Lee, and Toutanova}]{devlin2019bert}
Devlin, J.; Chang, M.; Lee, K.; and Toutanova, K. 2019.
\newblock {BERT:} Pre-training of Deep Bidirectional Transformers for Language Understanding.
\newblock In \emph{NAACL-HLT}, 4171--4186.

\bibitem[{Elbayad et~al.(2020)Elbayad, Gu, Grave, and Auli}]{elbayad2020depth}
Elbayad, M.; Gu, J.; Grave, E.; and Auli, M. 2020.
\newblock Depth-Adaptive Transformer.
\newblock In \emph{ICLR}.

\bibitem[{Fan, Grave, and Joulin(2020)}]{fan2020reducing}
Fan, A.; Grave, E.; and Joulin, A. 2020.
\newblock Reducing Transformer Depth on Demand with Structured Dropout.
\newblock In \emph{ICLR}.

\bibitem[{Geng et~al.(2021)Geng, Gao, Fu, and Zhang}]{geng2021romebert}
Geng, S.; Gao, P.; Fu, Z.; and Zhang, Y. 2021.
\newblock RomeBERT: Robust Training of Multi-Exit {BERT}.
\newblock \emph{CoRR}, abs/2101.09755.

\bibitem[{Goyal et~al.(2020)Goyal, Choudhury, Raje, Chakaravarthy, Sabharwal, and Verma}]{goyal2020power}
Goyal, S.; Choudhury, A.~R.; Raje, S.; Chakaravarthy, V.~T.; Sabharwal, Y.; and Verma, A. 2020.
\newblock PoWER-BERT: Accelerating {BERT} Inference via Progressive Word-vector Elimination.
\newblock In \emph{ICML}, 3690--3699.

\bibitem[{Guan et~al.(2022)Guan, Li, Leng, Lin, and Guo}]{guan2022transkimmer}
Guan, Y.; Li, Z.; Leng, J.; Lin, Z.; and Guo, M. 2022.
\newblock Transkimmer: Transformer Learns to Layer-wise Skim.
\newblock In Muresan, S.; Nakov, P.; and Villavicencio, A., eds., \emph{ACL}, 7275--7286.

\bibitem[{Hu et~al.(2022)Hu, Shen, Wallis, Allen{-}Zhu, Li, Wang, Wang, and Chen}]{edwardlora2021}
Hu, E.~J.; Shen, Y.; Wallis, P.; Allen{-}Zhu, Z.; Li, Y.; Wang, S.; Wang, L.; and Chen, W. 2022.
\newblock LoRA: Low-Rank Adaptation of Large Language Models.
\newblock In \emph{ICLR}.

\bibitem[{Jiao et~al.(2020)Jiao, Yin, Shang, Jiang, Chen, Li, Wang, and Liu}]{jiao2020tinybert}
Jiao, X.; Yin, Y.; Shang, L.; Jiang, X.; Chen, X.; Li, L.; Wang, F.; and Liu, Q. 2020.
\newblock TinyBERT: Distilling {BERT} for Natural Language Understanding.
\newblock In \emph{EMNLP}, Findings of EMNLP, 4163--4174.

\bibitem[{Kaya, Hong, and Dumitras(2019)}]{kaya2019shallow}
Kaya, Y.; Hong, S.; and Dumitras, T. 2019.
\newblock Shallow-deep networks: Understanding and mitigating network overthinking.
\newblock In \emph{ICML}, 3301--3310.

\bibitem[{Kim and Cho(2021)}]{kim2021length}
Kim, G.; and Cho, K. 2021.
\newblock Length-Adaptive Transformer: Train Once with Length Drop, Use Anytime with Search.
\newblock In \emph{ACL-IJCNLP}, 6501--6511.

\bibitem[{Kim et~al.(2021)Kim, Gholami, Yao, Mahoney, and Keutzer}]{kim2021ibert}
Kim, S.; Gholami, A.; Yao, Z.; Mahoney, M.~W.; and Keutzer, K. 2021.
\newblock {I-BERT:} Integer-only {BERT} Quantization.
\newblock In \emph{ICML}, 5506--5518.

\bibitem[{Kim et~al.(2022)Kim, Shen, Thorsley, Gholami, Kwon, Hassoun, and Keutzer}]{kim2022learned}
Kim, S.; Shen, S.; Thorsley, D.; Gholami, A.; Kwon, W.; Hassoun, J.; and Keutzer, K. 2022.
\newblock Learned Token Pruning for Transformers.
\newblock In \emph{SIGKDD}, 784--794.

\bibitem[{Kumar, Packer, and Koller(2010)}]{kumar2010self}
Kumar, M.; Packer, B.; and Koller, D. 2010.
\newblock Self-paced learning for latent variable models.
\newblock In \emph{NeurIPS}, 1189--1197.

\bibitem[{Li et~al.(2021)Li, Lin, Chen, Ren, Li, Zhou, and Sun}]{li2021cascadebert}
Li, L.; Lin, Y.; Chen, D.; Ren, S.; Li, P.; Zhou, J.; and Sun, X. 2021.
\newblock CascadeBERT: Accelerating Inference of Pre-trained Language Models via Calibrated Complete Models Cascade.
\newblock In \emph{EMNLP}, 475--486.

\bibitem[{Liao et~al.(2021)Liao, Zhang, Ren, Su, Sun, and He}]{liao2021pastfuture}
Liao, K.; Zhang, Y.; Ren, X.; Su, Q.; Sun, X.; and He, B. 2021.
\newblock A Global Past-Future Early Exit Method for Accelerating Inference of Pre-trained Language Models.
\newblock In \emph{NAACL-HLT}, 2013--2023.

\bibitem[{Lin(2004)}]{lin-2004-rouge}
Lin, C.-Y. 2004.
\newblock {ROUGE}: A Package for Automatic Evaluation of Summaries.
\newblock In \emph{Text Summarization Branches Out}, 74--81.

\bibitem[{Liu et~al.(2020)Liu, Zhou, Wang, Zhao, Deng, and Ju}]{liu2020fastbert}
Liu, W.; Zhou, P.; Wang, Z.; Zhao, Z.; Deng, H.; and Ju, Q. 2020.
\newblock FastBERT: a Self-distilling {BERT} with Adaptive Inference Time.
\newblock In \emph{ACL}, 6035--6044.

\bibitem[{Liu et~al.(2019)Liu, Ott, Goyal, Du, Joshi, Chen, Levy, Lewis, Zettlemoyer, and Stoyanov}]{liu2019roberta}
Liu, Y.; Ott, M.; Goyal, N.; Du, J.; Joshi, M.; Chen, D.; Levy, O.; Lewis, M.; Zettlemoyer, L.; and Stoyanov, V. 2019.
\newblock RoBERTa: {A} Robustly Optimized {BERT} Pretraining Approach.
\newblock \emph{CoRR}, abs/1907.11692.

\bibitem[{Michel, Levy, and Neubig(2019)}]{michel2019sixteen}
Michel, P.; Levy, O.; and Neubig, G. 2019.
\newblock Are sixteen heads really better than one?
\newblock In \emph{NeurIPS}, 14014--14024.

\bibitem[{Nallapati et~al.(2016)Nallapati, Zhou, dos Santos, G{\"{u}}l{\c{c}}ehre, and Xiang}]{ramesh2016cnn}
Nallapati, R.; Zhou, B.; dos Santos, C.~N.; G{\"{u}}l{\c{c}}ehre, {\c{C}}.; and Xiang, B. 2016.
\newblock Abstractive Text Summarization using Sequence-to-sequence RNNs and Beyond.
\newblock In Goldberg, Y.; and Riezler, S., eds., \emph{CoNLL}, 280--290.

\bibitem[{Ouyang et~al.(2022)Ouyang, Wu, Jiang, Almeida, Wainwright, Mishkin, Zhang, Agarwal, Slama, Ray et~al.}]{ouyang2022training}
Ouyang, L.; Wu, J.; Jiang, X.; Almeida, D.; Wainwright, C.; Mishkin, P.; Zhang, C.; Agarwal, S.; Slama, K.; Ray, A.; et~al. 2022.
\newblock Training language models to follow instructions with human feedback.
\newblock In \emph{NeurIPS}, volume~35, 27730--27744.

\bibitem[{Sanh et~al.(2019)Sanh, Debut, Chaumond, and Wolf}]{sanh2019distilBERT}
Sanh, V.; Debut, L.; Chaumond, J.; and Wolf, T. 2019.
\newblock DistilBERT, a distilled version of {BERT:} smaller, faster, cheaper and lighter.
\newblock \emph{CoRR}, abs/1910.01108.

\bibitem[{Schuster et~al.(2022)Schuster, Fisch, Gupta, Dehghani, Bahri, Tran, Tay, and Metzler}]{schuster2022confident}
Schuster, T.; Fisch, A.; Gupta, J.; Dehghani, M.; Bahri, D.; Tran, V.; Tay, Y.; and Metzler, D. 2022.
\newblock Confident Adaptive Language Modeling.
\newblock In \emph{NeurIPS}.

\bibitem[{Schuster et~al.(2021)Schuster, Fisch, Jaakkola, and Barzilay}]{schuster2021cat}
Schuster, T.; Fisch, A.; Jaakkola, T.~S.; and Barzilay, R. 2021.
\newblock Consistent Accelerated Inference via Confident Adaptive Transformers.
\newblock In \emph{EMNLP}, 4962--4979.

\bibitem[{Schwartz et~al.(2020)Schwartz, Stanovsky, Swayamdipta, Dodge, and Smith}]{schwartz2020right}
Schwartz, R.; Stanovsky, G.; Swayamdipta, S.; Dodge, J.; and Smith, N.~A. 2020.
\newblock The right tool for the job: Matching model and instance complexities.
\newblock In \emph{ACL}, 6640--6651.

\bibitem[{Sun et~al.(2019)Sun, Cheng, Gan, and Liu}]{sun2019patient}
Sun, S.; Cheng, Y.; Gan, Z.; and Liu, J. 2019.
\newblock Patient Knowledge Distillation for {BERT} Model Compression.
\newblock In \emph{EMNLP-IJCNLP}, 4322--4331.

\bibitem[{Sun et~al.(2022)Sun, Liu, Zhu, Geng, Wu, He, Ni, Xie, Huang, and Qiu}]{sun2022simple}
Sun, T.; Liu, X.; Zhu, W.; Geng, Z.; Wu, L.; He, Y.; Ni, Y.; Xie, G.; Huang, X.; and Qiu, X. 2022.
\newblock A simple hash-based early exiting approach for language understanding and generation.
\newblock In \emph{Findings of {ACL}}, 2409--2421.

\bibitem[{Taori et~al.(2023)Taori, Gulrajani, Zhang, Dubois, Li, Guestrin, Liang, and Hashimoto}]{taori2023alpaca}
Taori, R.; Gulrajani, I.; Zhang, T.; Dubois, Y.; Li, X.; Guestrin, C.; Liang, P.; and Hashimoto, T.~B. 2023.
\newblock Alpaca: A strong, replicable instruction-following model.
\newblock \emph{Stanford Center for Research on Foundation Models. https://crfm. stanford. edu/2023/03/13/alpaca. html}, 3(6): 7.

\bibitem[{Toneva et~al.(2019)Toneva, Sordoni, Combes, Trischler, Bengio, and Gordon}]{toneva2018empirical}
Toneva, M.; Sordoni, A.; Combes, R. T.~d.; Trischler, A.; Bengio, Y.; and Gordon, G.~J. 2019.
\newblock An empirical study of example forgetting during deep neural network learning.
\newblock In \emph{ICLR}.

\bibitem[{Touvron et~al.(2023)Touvron, Lavril, Izacard, Martinet, Lachaux, Lacroix, Rozi{\`{e}}re, Goyal, Hambro, Azhar, Rodriguez, Joulin, Grave, and Lample}]{hugo2023llama}
Touvron, H.; Lavril, T.; Izacard, G.; Martinet, X.; Lachaux, M.; Lacroix, T.; Rozi{\`{e}}re, B.; Goyal, N.; Hambro, E.; Azhar, F.; Rodriguez, A.; Joulin, A.; Grave, E.; and Lample, G. 2023.
\newblock LLaMA: Open and Efficient Foundation Language Models.
\newblock \emph{CoRR}, abs/2302.13971.

\bibitem[{Voita et~al.(2019)Voita, Talbot, Moiseev, Sennrich, and Titov}]{voita2019analyzing}
Voita, E.; Talbot, D.; Moiseev, F.; Sennrich, R.; and Titov, I. 2019.
\newblock Analyzing Multi-Head Self-Attention: Specialized Heads Do the Heavy Lifting, the Rest Can Be Pruned.
\newblock In \emph{ACL}, 5797--5808.

\bibitem[{Wang et~al.(2019)Wang, Singh, Michael, Hill, Levy, and Bowman}]{wang2018glue}
Wang, A.; Singh, A.; Michael, J.; Hill, F.; Levy, O.; and Bowman, S.~R. 2019.
\newblock GLUE: A multi-task benchmark and analysis platform for natural language understanding.
\newblock In \emph{ICLR}.

\bibitem[{Wang et~al.(2022)Wang, Chen, Chen, Shou, and McAuley}]{wang2022skipbert}
Wang, J.; Chen, K.; Chen, G.; Shou, L.; and McAuley, J. 2022.
\newblock Skipbert: Efficient inference with shallow layer skipping.
\newblock In \emph{ACL}, 7287--7301.

\bibitem[{Xiao et~al.(2023)Xiao, Lin, Seznec, Wu, Demouth, and Han}]{xiao2023smoothquant}
Xiao, G.; Lin, J.; Seznec, M.; Wu, H.; Demouth, J.; and Han, S. 2023.
\newblock Smoothquant: Accurate and efficient post-training quantization for large language models.
\newblock In \emph{ICML}, 38087--38099.

\bibitem[{Xin et~al.(2020)Xin, Tang, Lee, Yu, and Lin}]{xin2020deebert}
Xin, J.; Tang, R.; Lee, J.; Yu, Y.; and Lin, J. 2020.
\newblock DeeBERT: Dynamic early exiting for accelerating BERT inference.
\newblock In \emph{ACL}, 2246--2251.

\bibitem[{Xin et~al.(2021)Xin, Tang, Yu, and Lin}]{xin2021berxit}
Xin, J.; Tang, R.; Yu, Y.; and Lin, J. 2021.
\newblock {BER}xi{T}: Early Exiting for {BERT} with Better Fine-Tuning and Extension to Regression.
\newblock In \emph{EACL}, 91--104.

\bibitem[{Xu and McAuley(2023)}]{xu2023survey_dynamic}
Xu, C.; and McAuley, J.~J. 2023.
\newblock A Survey on Dynamic Neural Networks for Natural Language Processing.
\newblock In \emph{Findings of EACL}, 2325--2336.

\bibitem[{Xu et~al.(2015)Xu, Wang, Tian, Xu, Zhao, Wang, and Hao}]{xu2015stackoverflow}
Xu, J.; Wang, P.; Tian, G.; Xu, B.; Zhao, J.; Wang, F.; and Hao, H. 2015.
\newblock Short Text Clustering via Convolutional Neural Networks.
\newblock In \emph{NAACL-HLT}, 62--69.

\bibitem[{Yang et~al.(2019)Yang, Dai, Yang, Carbonell, Salakhutdinov, and Le}]{yang2019xlnet}
Yang, Z.; Dai, Z.; Yang, Y.; Carbonell, J.~G.; Salakhutdinov, R.; and Le, Q.~V. 2019.
\newblock XLNet: Generalized Autoregressive Pretraining for Language Understanding.
\newblock In \emph{NeurIPS}, 5754--5764.

\bibitem[{Yao et~al.(2022)Yao, Yazdani~Aminabadi, Zhang, Wu, Li, and He}]{yao2022zeroquant}
Yao, Z.; Yazdani~Aminabadi, R.; Zhang, M.; Wu, X.; Li, C.; and He, Y. 2022.
\newblock Zeroquant: Efficient and affordable post-training quantization for large-scale transformers.
\newblock In \emph{NeurIPS}, 27168--27183.

\bibitem[{Ye et~al.(2021)Ye, Lin, Huang, and Sun}]{ye2021trbert}
Ye, D.; Lin, Y.; Huang, Y.; and Sun, M. 2021.
\newblock {TR-BERT:} Dynamic Token Reduction for Accelerating {BERT} Inference.
\newblock In \emph{NAACL-HLT}, 5798--5809.

\bibitem[{Zhang et~al.(2019)Zhang, Kishore, Wu, Weinberger, and Artzi}]{zhang2019bertscore}
Zhang, T.; Kishore, V.; Wu, F.; Weinberger, K.~Q.; and Artzi, Y. 2019.
\newblock BERTScore: Evaluating Text Generation with BERT.
\newblock In \emph{ICLR}.

\bibitem[{Zhang et~al.(2022)Zhang, Zhu, Zhang, Wang, Jin, and Chung}]{zhang2022pcee}
Zhang, Z.; Zhu, W.; Zhang, J.; Wang, P.; Jin, R.; and Chung, T.-S. 2022.
\newblock Pcee-bert: Accelerating bert inference via patient and confident early exiting.
\newblock In \emph{Findings of NAACL}, 327--338.

\bibitem[{Zhou et~al.(2020)Zhou, Xu, Ge, McAuley, Xu, and Wei}]{zhou2020pabee}
Zhou, W.; Xu, C.; Ge, T.; McAuley, J.; Xu, K.; and Wei, F. 2020.
\newblock Bert loses patience: Fast and robust inference with early exit.
\newblock In \emph{NeurIPS}, volume~33, 18330--18341.

\bibitem[{Zhu(2021)}]{zhu2020leebert}
Zhu, W. 2021.
\newblock LeeBERT: Learned Early Exit for {BERT} with cross-level optimization.
\newblock In \emph{ACL-IJCNLP}, 2968--2980.

\end{thebibliography}
\section{Appendix}

\subsection{A: Loss Values}
In this part, We show more results about loss values on training and development sets on RTE, and MRPC datasets. 
In Figure \ref{fig:loss_distribution} (a) - (d) left, we found that the loss values under ConsistentEE are lower than that under the traditional loss, and the accuracy of each layer under the ConsistentEE objective is consistently higher than that under the traditional objective. 
This shows that a harsh training objective does not lead to an ambitious goal where all internal classifiers should predict all instances correctly. 
In Figure \ref{fig:loss_distribution} (a) - (d) right, we found that the pattern of the loss values and accuracy on the development set is similar to that on the training set. 
The accuracy of each layer under both objectives decreases on the development set. 
The accuracy under the traditional objective is stable after a specific layer. 
The accuracy under the ConsistentEE decreases after a specific layer because a ``hard'' instance is more likely to exit at a deeper layer. 
Deeper classifiers encounter many such ``hard'' instances.

\begin{figure*}[h]
 \centering
 
 \subfigure[Traditional Objectives on RTE dataset.]{\label{fig:rte_train}
  \includegraphics[width=0.7\textwidth]{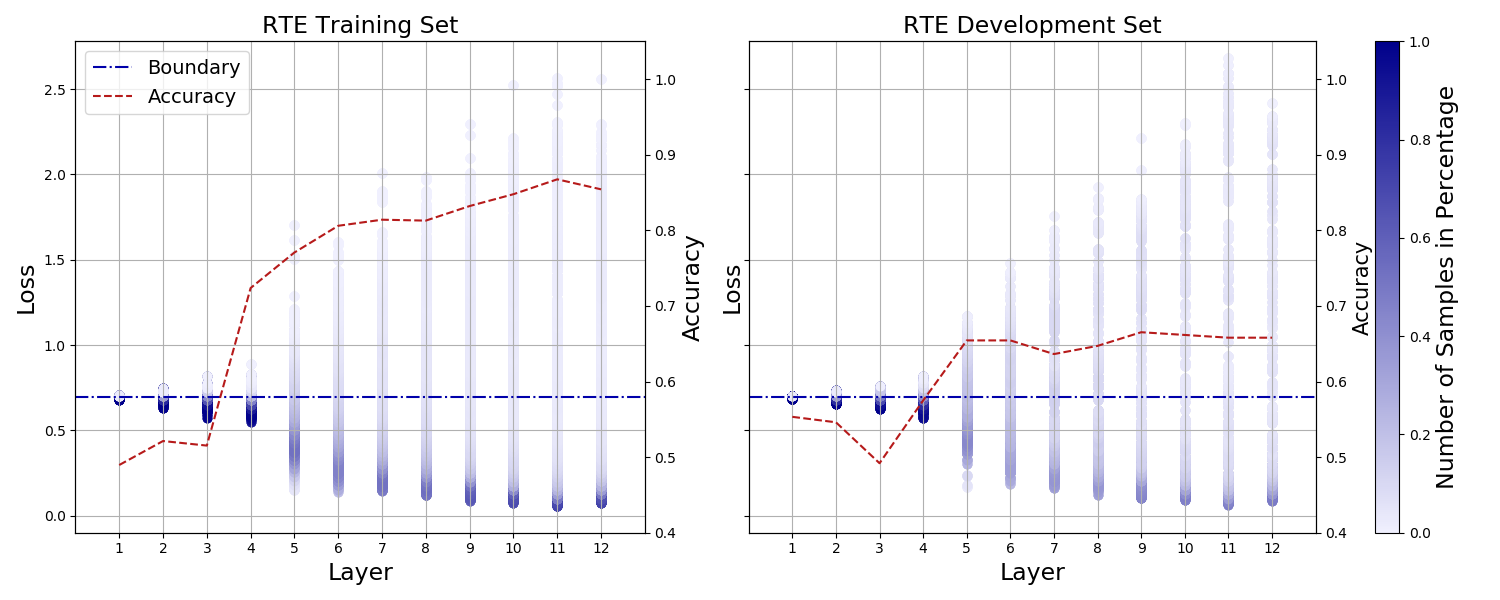}
 }
 
 \subfigure[ConsistentEE Objectives on RTE dataset.]{\label{fig:rte_eval}
  \includegraphics[width=0.7\textwidth]{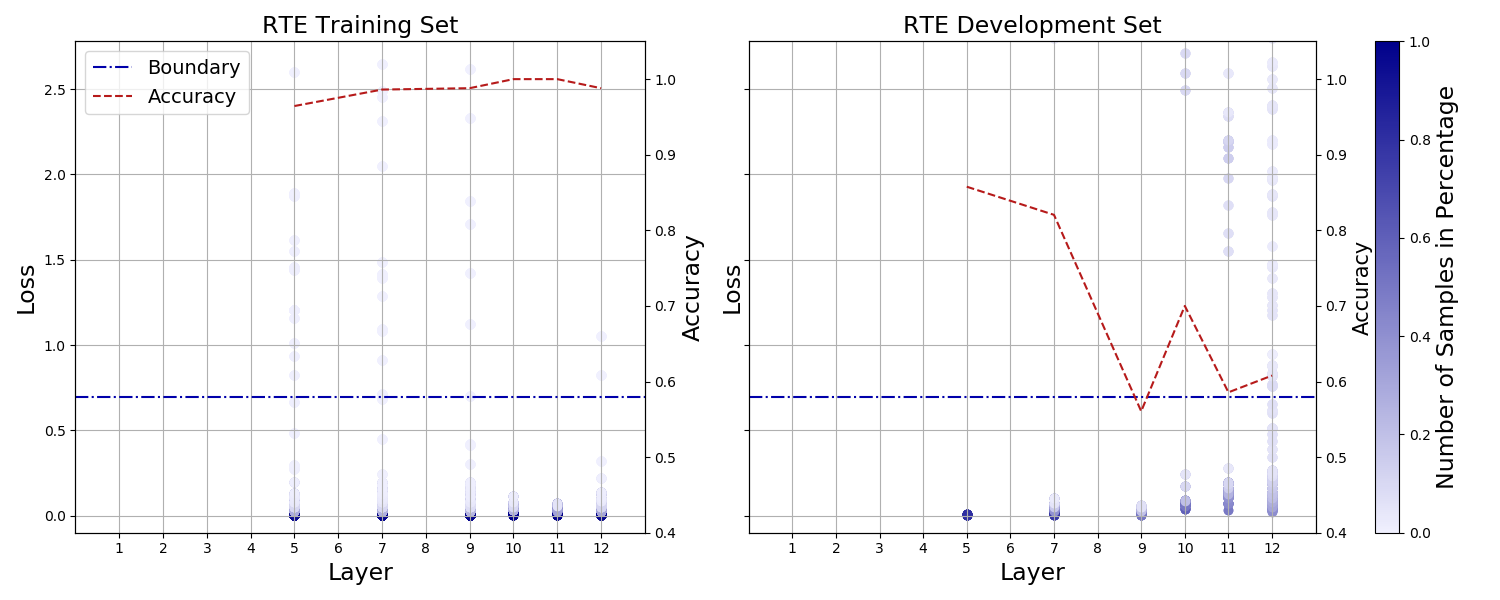}
 }

\subfigure[Traditional Objectives on MRPC dataset.]{\label{fig:mrpc_train}
  \includegraphics[width=0.7\textwidth]{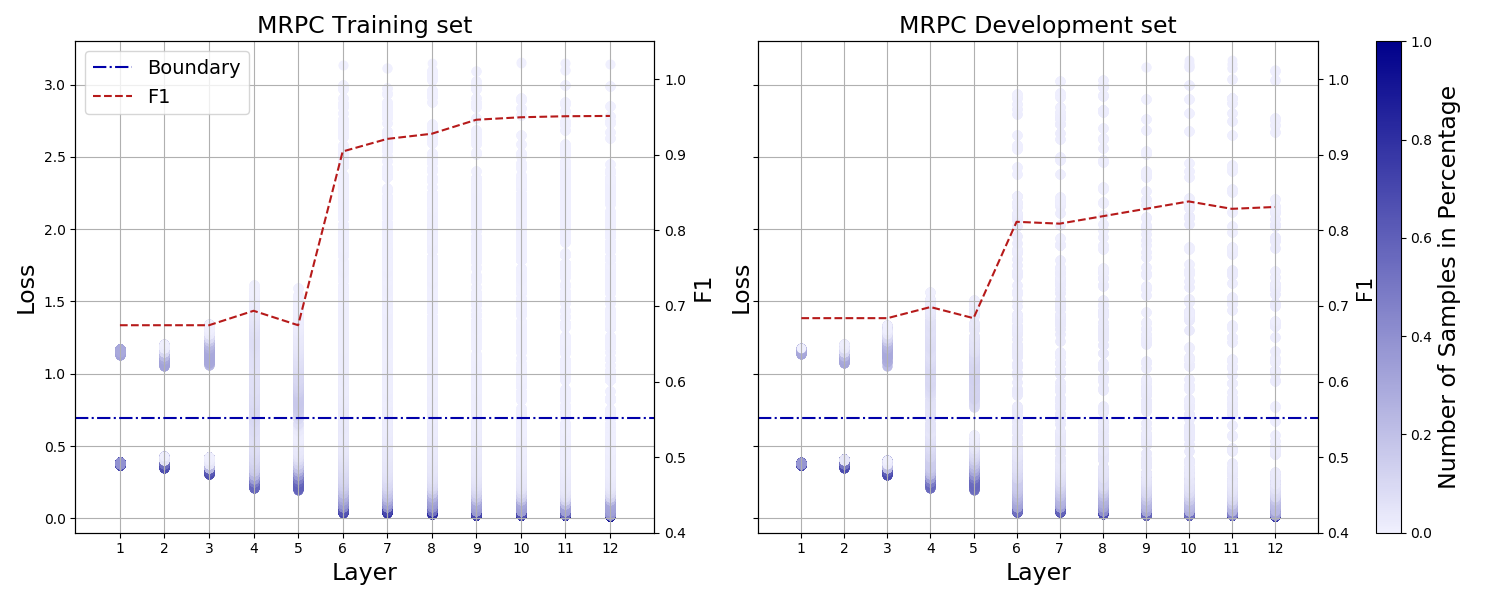}
 }

\subfigure[ConsistentEE Objectives on MRPC dataset.]{\label{fig:mrpc_eval}
  \includegraphics[width=0.7\textwidth]{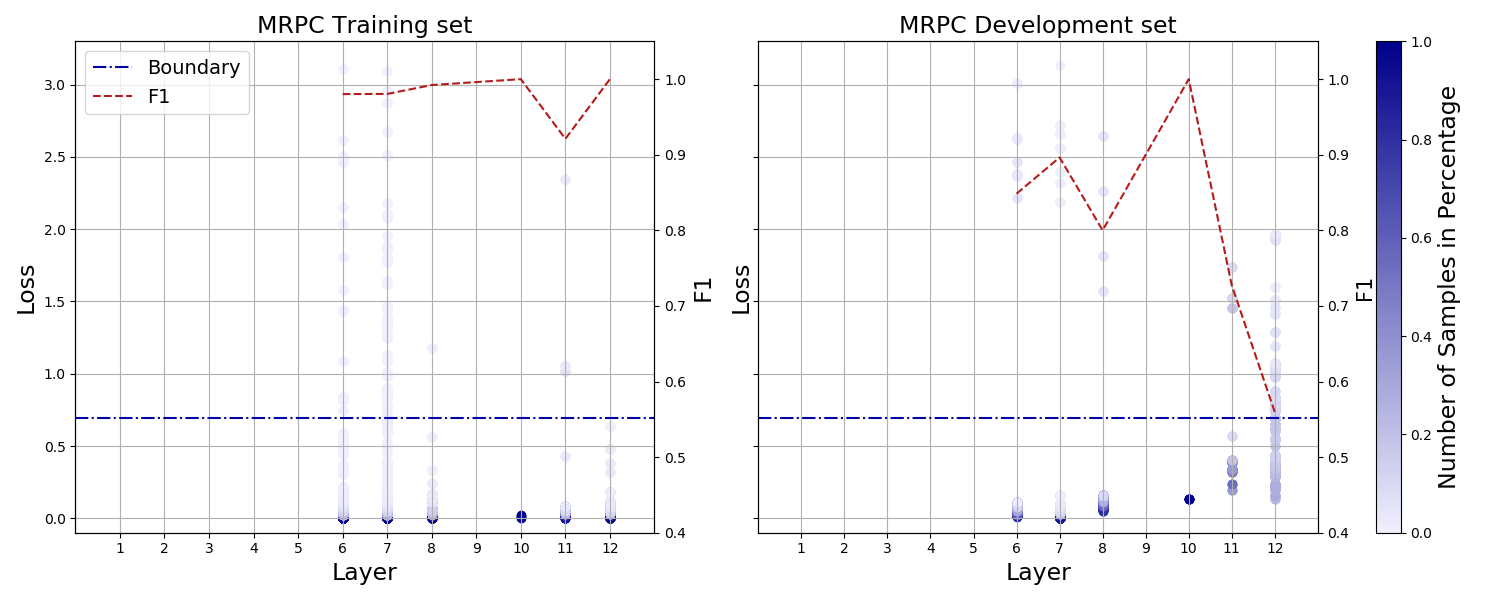}
}

 \vspace{-0.1in}
 \caption{Loss values at different layers on training and development dataset using the weighted sum objective and ConsistentEE objective respectively. The dashed dot line is the classification boundary. A loss above the boundary means misclassification.  
 The dashed line is the classification accuracy of each layer. The darker the color, the more samples share the same loss value.  
 } \label{fig:loss_distribution}
\end{figure*}

\subsection{B: Results on different PLM backbones:}

We provide more results on different PLM backbones including ALBERT-Base, BERT-Large, ALBERT-Large in Table \ref{tab:plm_main_result}. 
Our method consistently outperforms baselines.

\begin{table}[H]
	% \centering
	\scalebox{0.75}{
% 		\begin{tabular}{l|p{0.8cm}<{\centering} p{0.82cm}<{\centering} p{0.8cm}<{\centering} p{0.82cm}<{\centering} p{0.8cm}<{\centering} p{0.8cm}<{\centering} p{0.8cm} }
\begin{tabular}{l|*{6}{p{0.91cm}<{\centering}}}

			\toprule
			Method	 & \multicolumn{2}{c|}{RTE} & \multicolumn{2}{c|}{MRPC} & \multicolumn{2}{c}{SST-2}  \\	
   
    &  Acc & Layer  &  F1 & Layer  &   Acc & Layer  \\
			\midrule 
			 
            % \multicolumn{19}{c}{\textbf{Test Set}}\\
                % \midrule
                % \textbf{ALBERT-Base}  & 75.1(75.1) & 12 & 86.7(89.8) & 12 & 94.1(91.5) & 12 
                % \\
                % PABEE   & 74.4  & -10 \% &  86.3 & -48\% &93.8 &-37\%   \\
                % PCEE-BERT  & 74.7  & -18\% &  86.5(89.6)  & -53\% &  94.0& -48\% 
                % \\
                
                \textbf{ALBERT-Base}  & 75.1 & 12 & 86.7 & 12 & 94.1 & 12 
                \\
                %PABEE   & 74.4  & -10\% &  86.3 & -48\% &93.7 &-37\%   \\ original data
                PABEE   & 74.4  & -10\% &  86.3 & -28\% &93.7 &-37\%   \\
                %PCEE-BERT  & 74.7  & -18\% &  86.5 & -53\% &  93.9 & -48\% \\ original data
                PCEE-BERT  & 74.7  & -18\% &  86.5 & -23\% &  93.9 & -38\%  \\
                
                \textbf{ConsistentEE}
			 & \textbf{75.0} & \textbf{ -25\%} & \textbf{86.7} & \textbf{-34\%} & \textbf{94.0} & \textbf{-43\%} \\  

               \midrule 
               
               \textbf{BERT-Large}  & 70.1 & 24 & 89.3 & 24 & 94.8 & 24 
                \\
                %PABEE   & 69.0 & -19\% &  86.6  & -33\% &   93.8 & -42\%   \\ %rte: 66.8 -36% original data
                PABEE   & 69.0 & -19\% &  86.6  & -13\% &   93.8 & -22\%   \\ 
                %PCEE-BERT  &  69.3 & -37\% & 88.9   & -32\% & 93.8 & -25\% \\ original data
                PCEE-BERT  &  69.3 & -17\% & 88.9   & -12\% & 93.8 & -25\% \\
                \textbf{ConsistentEE}
			 & \textbf{69.5 } & \textbf{ -26\%} & \textbf{89.2 } & \textbf{ -22\%} & \textbf{94.1 } & \textbf{-30\%} \\

            \midrule
            \textbf{ALBERT-Large}  & 83.1 & 24 & 92.0 & 24 & 94.9 & 24 
                \\
                %PABEE   & 81.2  & -25\% &  91.5  &  -36\% &  \textbf{94.8} & -42\%   \\ original data
                PABEE   & 81.2  & -15\% &  91.5  &  -26\% &  \textbf{94.8} & -22\%   \\
                %PCEE-BERT  & 82.2  &  -28\% &  91.6  & -46\% &94.5 &   -37\%    original data
                PCEE-BERT  & 82.2  &  -18\% &  91.6  & -36\% &94.5 &-27\%
                \\
                \textbf{ConsistentEE}
			 & \textbf{82.5} & \textbf{-24\%} & \textbf{91.8} & \textbf{ -39\%} & \textbf{94.8} & \textbf{ -35\%} \\ 
                
			\bottomrule
    	\end{tabular}}
 	\caption{F1 / Accuracy on RTE, MRPC, SST-2 test sets.} 
  %Comparison on three GLUE datasets  using different backbones including ALBERT-Base, BERT-Large and ALBERT-Large. 
%The evaluation metrics for model quality are accuracy or F1 scores. The evaluation metrics for model acceleration are relative saved layers/runtime (w.r.t BERT-Base). Results which exhibit better accuracy and saved layers than other methods are highlighted in \textbf{bold}.
    % Averaged scores and saved layers/runtime on eight datasets are reported. 
    \label{tab:plm_main_result}
\end{table}

\subsection{C: Case study on easy and hard examples}

\textcolor{blue}{Examples on Classification task (SST-2) on the backbone of BERT-base (12 layers): }

\textbf{Easy} (exiting at \textbf{2nd} layer): (1) ``it 's a charming and often affecting journey;'' (2)``i had to look away - this was god awful;'' 

\textbf{Hard} (exiting at \textbf{12th} layer): (1)``pumpkin takes an admirable look at the hypocrisy of political correctness , but it does so with such an uneven tone that you never know when humor ends and tragedy begins; '' (2)``this is the sort of burly action flick where one coincidence pummels another , narrative necessity is a drunken roundhouse , and whatever passes for logic is a factor of the last plot device left standing . ''

\noindent \textcolor{blue}{Examples on Generation task (Alpaca/Dolly) on the backbone of LLaMA-7b (32 layers):} 

\textbf{Easy} (averagely exiting at \textbf{7.73th} layer): (1) Instruction: What is Python? Response: A computer programming language that can impart APIs, third-party apps, and web content. (2) Instruction: Why mobile is bad for human? Response: The use of mobile phones has been found to have a negative impact on human well-being. The tracking of human activities through mobile phone usage has been shown to reduce cognitive processes and increase the level of social limitations. 

\textbf{Hard} (averagely exiting at \textbf{31.17th} layer): (1) Instruction: What is a dispersive prism? Response: A dispersive prism is an optical prism that is used to disperse light, that is, to separate light into its spectral components (the colors of the rainbow). Different wavelengths (colors) of light will be deflected by the prism at different angles. (2) Instruction: write a scene between two actors discussing movie Inception. Response: Actor 1: "Have you seen Inception?"
Actor 2: "Yes, I have. It was an amazing movie."
Actor 1: "I know, right? I was so impressed with the visuals and the storyline."
Actor 2: "Yeah, I agree. It was a great movie."
Actor 1: "I'm so glad I saw it. It was so thought-provoking."
Actor 2: "I know, it really made me think about the concept of dreams and reality."

\subsection{D: Examples of generation from ConsistentEE}

Examples of generation from ConsistentEE under different speedup ratios result on Alpaca/Dolly are shown in Table \ref{tab:case_study} and Table \ref{tab:case_study_CNNDM}. 
The responses generated under 2.0x appear reasonable and closely resemble the original response. However, the responses generated under 5.0x vary from the original response but still make sense. 

\begin{table*}[p]
	\centering
	\scalebox{0.88}{
		\begin{tabular}{p{4cm}p{5cm}p{5cm}p{5cm}} 
			\toprule
			Instruction & Original Response & Response 2x & Response 5x \\

           		    \midrule          
            What is best? Playstation or Xbox        &  Playstation and Xbox are two competing game consoles, where which is ``best'' will depend on the preferences of the gamer. &I think Playstation is better because it has more games and it is more fun. &I prefer Playstation because Playstation has more games and more games are better. \\
               \midrule 
                What happens when the sun goes down?
               & The sun goes down and the moon comes up. & There are darknesses all around us when sunlight vanishes.& There are night lights lit up.
             \\ 
             \midrule
                         How should I prioritize my time??
                    &Prioritize work that is tied to specific goals and projects.  & Prioritize your time by investigating the needs of your daily life and focusing on the most crucial tasks. & Prioritize work that has a meaningful impact on your goals or vision.    \\
			\bottomrule
   	\end{tabular}}
   \caption{Snapshot of model responses under different speedup ratios on Alpaca/Dolly. 
   }
   \label{tab:case_study}
\end{table*}

\begin{table*}[p]
	\centering
	\scalebox{0.78}{
		\begin{tabular}{p{9cm}p{4cm}p{4cm}p{4cm}} 
			\toprule
			Context & Original Response & Response 2x & Response 5x \\		
           		    \midrule          
            Five Americans who were monitored for three weeks at an Omaha, Nebraska, hospital after being exposed to Ebola in West Africa have been released, a Nebraska Medicine spokesman said in an email Wednesday. One of the five had a heart-related issue on Saturday and has been discharged but hasn't left the area, Taylor Wilson wrote. The others have already gone home. They were exposed to Ebola in Sierra Leone in March, but none developed the deadly virus. They are clinicians for Partners in Health, a Boston-based aid group...&  Five Americans who were exposed to Ebola in West Africa were released from a Nebraska hospital after three weeks of monitoring, with one still in the area for heart-related issue. The CDC expects to release the last of 17 patients by Thursday. & Five Americans exposed to Ebola in West Africa have been released from a Nebraska hospital after monitoring for three weeks. One has a heart-related issue and remains in the area, while the others have gone home. &Five Americans who were exposed to Ebola in Sierra Leone were released from a Nebraska hospital after three weeks of monitoring, with one still in the area for further evaluation. \\
               \midrule 
               (CNN)For the first time in eight years, a TV legend returned to doing what he does best. Contestants told to "come on down!" on the April 1 edition of "The Price Is Right" encountered not host Drew Carey but another familiar face in charge of the proceedings. Instead, there was Bob Barker, who hosted the TV game show for 35 years before stepping down in 2007...
                &Bob Barker returned to host "The Price Is Right" for the first time in 8 years, filling in for Drew Carey. At 91, he looked spry and easily handled hosting duties, turning over to Carey at the end.& Bob Barker, former host of "The Price is Right" returned to the show after 8 years, surprising contestants and hosting the first game with Drew Carey. & Bob Barker returned to hosting "The Price is Right" for the first time in 8 years, filling in for Drew Carey during a taping of the show.
             \\ 
			\bottomrule
   	\end{tabular}}
   \caption{Snapshot of model responses under different speedup ratios on CNN-DM. 
   }
   \label{tab:case_study_CNNDM}
\end{table*}

\end{document}